\documentclass[10pt,twocolumn,letterpaper]{article}

\usepackage[pagenumbers]{cvpr} 

\definecolor{cvprblue}{rgb}{0.21,0.49,0.74}
\definecolor{c846358}{HTML}{846358}
\definecolor{c0c8599}{HTML}{0c8599}
\usepackage[pagebackref,breaklinks,colorlinks,allcolors=cvprblue]{hyperref}
\usepackage{newtxtext,newtxmath}
\usepackage{multirow}
\usepackage[table,xcdraw]{xcolor}
\usepackage{algorithm}    
\usepackage{algpseudocode}  
\usepackage{amsmath}     
\usepackage{amssymb}     
\usepackage{booktabs}
\usepackage{multirow}
\usepackage{graphicx}
\usepackage{enumitem}
\usepackage{adjustbox}
\def\paperID{8567} 
\def\confName{CVPR}
\def\confYear{2026}

\title{Dejavu: Towards Experience Feedback Learning for Embodied Intelligence}

\author{Shaokai Wu, Yanbiao Ji, Qiuchang Li, Zhiyi Zhang, Qichen He,\\
Wenyuan Xie, Guodong Zhang, Bayram Bayramli, Yue Ding\textsuperscript{\rm *}, Hongtao Lu\textsuperscript{\rm *} \\
School of Computer Science, Shanghai Jiao Tong University
}

\begin{document}
\maketitle

\begin{abstract}
Embodied agents face a fundamental limitation: once deployed in real-world environments, they cannot easily acquire new knowledge to improve task performance. In this paper, we propose Dejavu, a general post-deployment learning framework that augments a frozen Vision-Language-Action (VLA) policy with retrieved execution memories through an Experience Feedback Network (EFN). EFN identifies contextually relevant prior action experiences and conditions action prediction on the retrieved guidance. We train EFN with reinforcement learning and semantic similarity rewards, encouraging the predicted actions to align with past behaviors under the current observation. During deployment, EFN continually expands its memory with new trajectories, enabling the agent to exhibit ``learning from experience.'' Experiments across diverse embodied tasks show that EFN improves adaptability, robustness, and success rates over frozen baselines. Our Project Page is  \href{https://dejavu2025.github.io/}{https://dejavu2025.github.io/}.
\end{abstract}

\let\thefootnote\relax    
\footnotetext[0]{\textsuperscript{*}Corresponding authors.}
\section{Introduction}
Embodied intelligence studies agents that learn and act through physical interaction with their environment \cite{sun2024comprehensive, liu2025embodied}. Recently, unified Vision–Language–Action (VLA) models have shown impressive generalization across diverse tasks \cite{zitkovich2023rt,shao2025large,firoozi2025foundation,han2025multimodal}, but only after massive offline training on fixed data distributions \cite{Brohan2023RT1,team2024octo}. Once deployed, their weights, and thus their knowledge, remain fixed unless the model is retrained \cite{liu2025robot}, so most real systems effectively ``stop learning'' in the wild \cite{liu2025robot}. 

This raises a natural question: must intelligent systems always rewrite their internal weights to improve? Humans often tackle new problems by recalling and reusing past experiences rather than acquiring entirely new facts \cite{andrychowicz2017hindsight,oh2018self}. Episodic memory, which underlies the feeling of ``déjà vu,'' does not alter core knowledge representations but does enable rapid adaptation by analogy \cite{goyal2022retrieval}. We therefore ask whether an AI agent can similarly improve by recalling and reusing its own ``experiences.'' If a neural network could leverage stored memories of situations and solutions to inform its current decisions, it could improve over time at inference simply by accumulating and exploiting new experiences, without gradient-based retraining~\cite{blundell2016model,pritzel2017neural,kuznetsov2021solving}. This idea of ``learning from déjà vu'' motivates our approach.

A closely related literature studies retrieval-augmented reinforcement learning and retrieval-augmented embodied agents. Retrieval-augmented~\cite{lewis2020retrieval} RL couples an external experience buffer with a parametric policy or value function, retrieving past trajectories similar to the current state to improve credit assignment or value estimation \cite{goyal2022retrieval}. In embodied and robotic control, recent methods retrieve prior executions in various forms, such as state--action trajectories or policy snippets, and leverage them during decision making, either by fine-tuning on retrieved data \cite{zhu2024retrieval,memmel2024strap} or by conditioning the policy directly on retrieved trajectories as additional context \cite{kuroki2024multi,kuang2024ram,kwon2025rt}. These works share a retrieval-plus-decision paradigm \cite{goyal2022retrieval,zhu2024retrieval,memmel2024strap}, but they typically target a trainable policy whose weights continue to change during or after deployment \cite{kuang2024ram,kuroki2024multi}, operate on static offline corpora rather than live memories that grow during deployment \cite{goyal2022retrieval,zhu2024retrieval}, and define retrieval over compact state or task abstractions instead of the rich, open-vocabulary vision--language interface of modern VLAs \cite{zhu2024retrieval,kwon2025rt}. As a result, existing retrieval-augmented methods do not offer a simple deployment-time mechanism for continually improving a frozen, unified policy. To our knowledge, EFN is the first post-deployment learning framework for VLA policies that couples a live experience bank with retrieval-conditioned residual correction, enabling improvement through memory growth without backbone finetuning.

To this end, we propose the \emph{Experience Feedback Network} (EFN), an experience-centric controller for improving pre-trained VLA policies at deployment without modifying their weights. EFN takes the current observation together with an action retrieved from a matched experience and predicts a residual action, which is added to the VLA output to produce the final control. When a strong prior experience exists, EFN learns to exploit it to refine the action; when retrieval is uninformative, it falls back toward the base policy. We train EFN with reinforcement learning using a dense similarity-based reward: the agent receives higher reward when the next observation resembles the \emph{next observation in the retrieved experience}, providing frequent shaping signals beyond sparse success--failure feedback. Concretely, EFN learns retrieval-conditioned residual control by treating the successor of a retrieved transition as a semantic target for shaping. EFN is optimized with Soft Actor--Critic~\cite{haarnoja2018soft}, whose entropy regularization and off-policy sample efficiency facilitate effective reuse of stored experiences. 

During deployment, we maintain a \emph{live} experience bank that is continuously augmented with newly successful rollouts. At each inference step, the agent retrieves a similar trajectory in a joint vision--language embedding space, provides the matched experience action to EFN alongside the current observation, and obtains a residual that refines the base VLA output. This yields a retrieval-augmented yet \emph{weight-frozen} VLA agent that improves purely by accumulating and reusing its own experiences over time. We integrate EFN with OpenVLA~\cite{kim2024openvla}, UniVLA~\cite{bu2025univla}, and GO-1~\cite{bu2025agibot}, and evaluate it in both simulation and on a real-world platform. Extensive experiments show that EFN not only improves success and efficiency over strong VLA baselines, but also outperforms representative retrieval-augmented RL and embodied baselines under the same experience bank and evaluation protocol. 

\noindent We summarize our main contributions as follows:
\begin{itemize}[leftmargin=*,topsep=0pt]
  \setlength\itemsep{0mm}
  \item We propose EFN, to our knowledge the first post-deployment learning framework for \emph{frozen} VLA policies that improves behavior through online experience accumulation and retrieval-conditioned residual correction, enabling adaptation by memory updates rather than backbone finetuning.
  \item We design a live experience bank that stores synchronized vision--language--action trajectories, retrieves task-relevant transitions in a joint embedding space, and trains the residual with similarity-shaped reinforcement signals toward retrieved successors.
  \item We integrate EFN with OpenVLA, UniVLA, and GO-1, and benchmark it on LIBERO and a real-world AgiBot-G1 setup, showing consistent gains over retrieval-only, residual-only, retrieval-augmented RL, and test-time training baselines under the same interaction budget.
\end{itemize}

\begin{figure*}[t]
 \centering
 \includegraphics[width=0.995\textwidth]{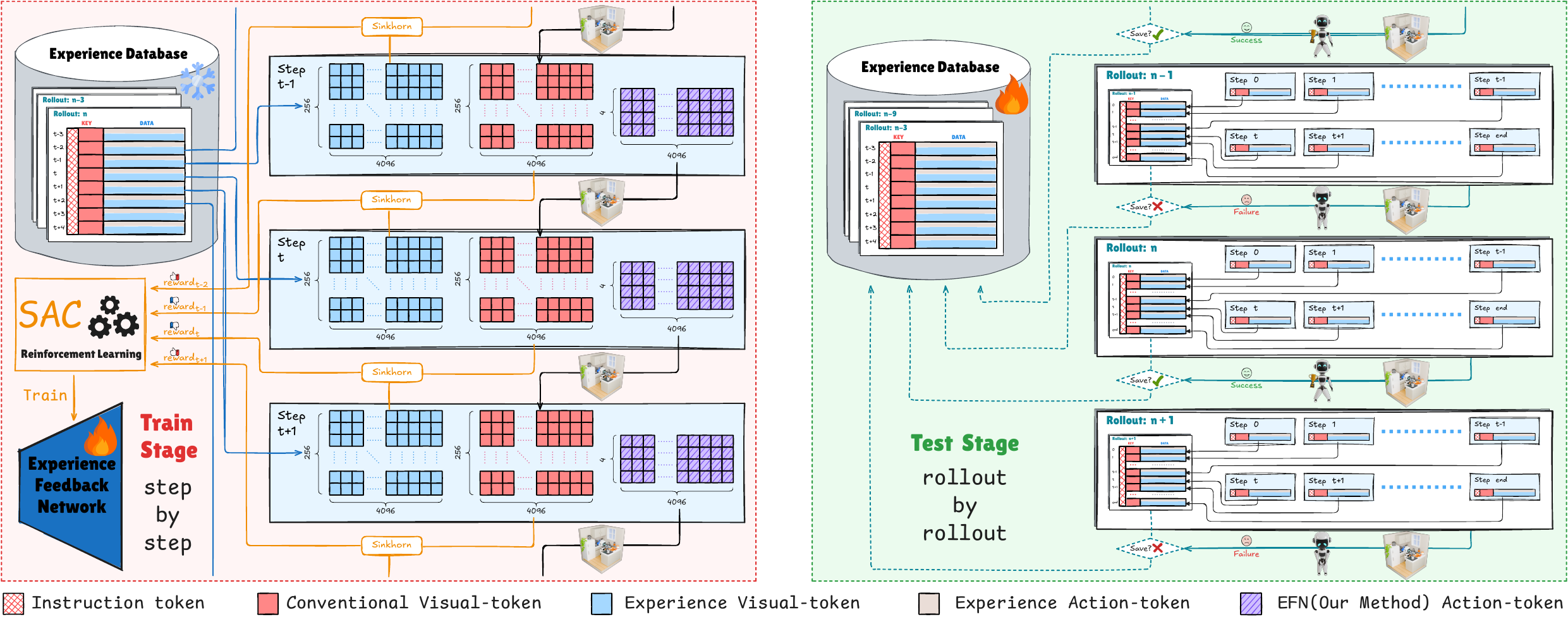}
 \vspace{0.3em}
 \begin{minipage}{0.48\textwidth}
  \captionof{figure}{EFN trains a residual policy with SAC to nudge
  the base action so the next frame matches the successor in memory.}
  \label{fig:efn-train}
 \end{minipage}\hfill
 \begin{minipage}{0.48\textwidth}
  \captionof{figure}{EFN performs inference by retrieving
  candidates, applying residual correction, and growing the experience bank online.}
  \label{fig:efn-test}
 \end{minipage}
 \vspace{-10mm}
\end{figure*}

\section{Background}
\label{sec:related}
\subsection{Vision-Language-Action (VLA) Models}
Large-scale VLA policies couple open-vocabulary perception with end-to-end control and achieve strong cross-task generalization in manipulation~\cite{shukor2025smolvla,zhang2025pure,zhai2025igniting,cheang2025gr,Brohan2023RT1,zitkovich2023rt}. Recent open-source generalist policies~\cite{team2024octo,kim2024openvla} extend this paradigm to heterogeneous robots and sensors, while lighter-weight architectures improve inference efficiency for deployment~\cite{liu2024robomamba}. Benchmarks such as LIBERO~\cite{liu2023libero} provide a standardized testbed for compositional generalization and post-deployment robustness.

In this work, we adopt the emerging view of VLAs as \emph{frozen backbones}: the VLA provides a strong yet imperfect policy, and EFN operates purely as an external correction module without modifying the backbone weights.

\subsection{Post-deployment Learning and Retrieval}
A central deployment challenge is improving behavior \emph{after} a policy has been trained, ideally without repeatedly finetuning large models. Human-in-the-loop and continual-learning frameworks study how robots can collect feedback and adapt online~\cite{liu2025robot,liu2023libero}. In parallel, retrieval-augmented reinforcement learning associates policies or value functions with an external experience buffer and retrieves similar trajectories at test time to guide decision making~\cite{goyal2022retrieval,toteja2025context,bacciu2023rraml,tarasov2025yes}. In robotics, retrieval has been used to condition policies on past episodes, demonstrations, or cached behaviors for training-free guidance.

Residual policy learning improves a strong controller by predicting an additive correction rather than a full action, enabling safer and more sample-efficient finetuning~\cite{johannink2018residual, silver2018residual}. Dense, similarity-shaped rewards and world models~\cite{hafner2025mastering,bagatella2025test} further stabilize RL by comparing predicted or observed futures to goal-like targets instead of relying solely on sparse task success.

\textbf{Our perspective.}
EFN lies at the intersection of these ideas. It targets the post-deployment regime with a \emph{frozen} VLA, maintains a live experience bank, retrieves a task-relevant transition in a joint vision--language space, and predicts a residual action that refines the base policy. The residual is trained with dense similarity-shaped rewards that compare the observed next frame to the retrieved successor, combining retrieval-conditioned guidance with residual correction without ever updating the large backbone.

\subsection{Embodied Reinforcement Learning}
Embodied RL trains control policies via interaction with physical or simulated environments, often under sparse rewards and limited data. Prior work addresses these issues through goal relabeling~\cite{andrychowicz2017hindsight}, large-scale real-robot data collection~\cite{bodnar2019quantile}, leveraging offline datasets before online improvement~\cite{nair2020awac}, strong regularization from pixels~\cite{yarats2021mastering}, and world-model--based imagination~\cite{hafner2025mastering}. Benchmarks for compositional and continual embodied learning further emphasize evaluating robustness and knowledge accumulation~\cite{liu2023libero,yang2025embodiedbench,zhang2025vlabench,garcia2025towards}.

\textbf{Our scope.}
Rather than training ever larger VLAs, we focus on a lightweight post-deployment mechanism that augments a frozen policy. Concretely, EFN is an Experience Feedback Network trained with Soft Actor--Critic~\cite{haarnoja2018soft} to map retrieved trajectories and current observations to residual actions that refine the controller, effectively \emph{learning an experience-feedback module} instead of relearning the VLA.

\section{Methodology}
\subsection{Overview of EFN}
EFN wraps a frozen vision--language--action (VLA) policy with an experience-centric controller that improves behavior at deployment \emph{without} updating the backbone weights. 
%
We first design a step-level experience bank and storage schema that record both rollout-level instructions and per-step visual keys and actions (Sec.~\ref{sec:experience-bank}). Given this bank, EFN learns a residual policy on top of the frozen VLA using a similarity-shaped reward and Soft Actor--Critic, with additional shaping to discourage idling (Sec.~\ref{sec:residual-policy}). At deployment, all network parameters remain frozen: EFN performs instruction-filtered, efficiency-aware retrieval, applies the trained residual correction at each step, and grows the bank online by appending successful rollouts (Sec.~\ref{sec:deployment}). Post-deployment adaptation therefore arises purely from memory growth and recall, not from further gradient updates. Figure~\ref{fig:efn-train} and Figure~\ref{fig:efn-test} summarize EFN in the training and deployment pipelines, respectively. We elaborate on the rationale behind this design in Appendix Section\ref{rationale}.

\begin{figure}[t]
 \centering
 \begin{subfigure}[t]{0.495\linewidth}
  \centering
  \includegraphics[height=3.3cm,keepaspectratio]{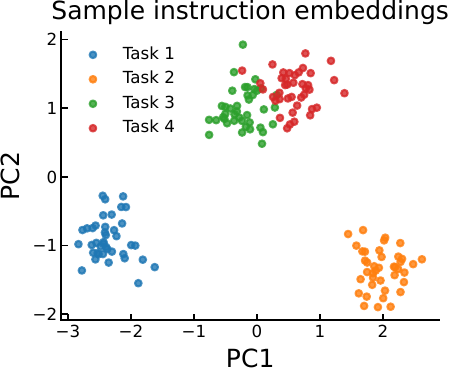}
  \caption{Instruction embeddings.}
  \label{fig:efn-retrieval:a}
 \end{subfigure}\hfill
 \begin{subfigure}[t]{0.495\linewidth}
  \centering
  \includegraphics[height=3.3cm,keepaspectratio]{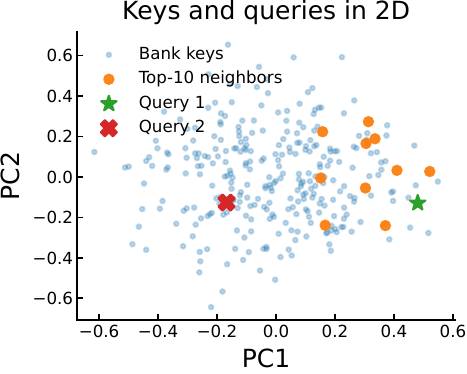}
  \caption{Keys and queries.}
  \label{fig:efn-retrieval:b}
 \end{subfigure}
 \caption{
 Visualization of EFN's language-conditioned retrieval.
 (a) PCA projection of instruction embeddings $\ell_\tau$ grouped by task type:
 similar instructions form clusters, allowing EFN to restrict retrieval to a small set of relevant rollouts.
 (b) PCA projection of experience keys $\mathbf{k}_i$ and online queries $\mathbf{q}_t$:
 for a given query, the retrieved top-$k$ neighbors (blue) lie nearby in this space.
 }
 \vspace{-8pt}
 \label{fig:efn-retrieval}
\end{figure}

\subsection{Experience Bank Design}
\label{sec:experience-bank}
\noindent \textbf{Storage schema.}
We organize experience as full \emph{rollouts} $\tau=(s_1,a_1,\dots,s_T,a_T)$ and insert into the bank every step $(s_t,a_t)$ encountered during data collection or deployment, i.e., every step at which the robot executes a control action. We exclude the initial ``blank'' waiting steps during simulator loading and robot warm-up from training and evaluation.

The initial offline bank contains both successful and near-successful trajectories, and occasionally failed rollouts, which we retain because our reward is defined on semantic similarity rather than a binary success flag; see Appendix Section\ref{sec:success-failure} for discussion. During deployment, however, we append only trajectories that reach the goal, using these successful rollouts as high-quality references.

For each rollout $\tau$, we also store a fixed \emph{instruction embedding} $\ell_\tau$, obtained by encoding the task description with the VLA's language model at the beginning of the episode. At the step level, we record three items: (i) the VLA vision-encoder features $\mathbf{F}_t \in \mathbb{R}^{L\times C}$ for frame $s_t$ (e.g., token features), (ii) a compact \emph{key} vector $\mathbf{k}_t \in \mathbb{R}^{d_k}$ derived from $\mathbf{F}_t$ for retrieval, and (iii) the base policy's raw action $\mathbf{a}_t^{(0)}$ executed at that step. The bank therefore stores tuples $\big(\ell_\tau,\mathbf{F}_t,\mathbf{k}_t,\mathbf{a}_t^{(0)}\big)$ for all valid $t$ across all trajectories.

\noindent \textbf{Key construction and probabilistic top-$k$ retrieval.}
We form retrieval keys using mean–max fusion with per-token $\ell_2$ normalization. First, $\ell_2$-normalize each token feature in $\mathbf{F}_t$ across channels, then compute mean and max along the token dimension and normalize each result:
\begin{equation}
\begin{aligned}
\tilde{\mathbf{F}}_t(\ell,\cdot) &= 
  \frac{\mathbf{F}_t(\ell,\cdot)}
     {\left\lVert \mathbf{F}_t(\ell,\cdot)\right\rVert_2+\varepsilon}, \\[3pt]
\mathbf{m}_t &= 
  \frac{\mathrm{mean}_\ell\!\left(\tilde{\mathbf{F}}_t(\ell,\cdot)\right)}
     {\left\lVert \mathrm{mean}_\ell\!\left(\tilde{\mathbf{F}}_t(\ell,\cdot)\right)\right\rVert_2+\varepsilon}, \\[3pt]
\mathbf{x}_t &= 
  \frac{\mathrm{max}_\ell\!\left(\tilde{\mathbf{F}}_t(\ell,\cdot)\right)}
     {\left\lVert \mathrm{max}_\ell\!\left(\tilde{\mathbf{F}}_t(\ell,\cdot)\right)\right\rVert_2+\varepsilon}.
\end{aligned}
\end{equation}
We then fuse the two by an equal-weight average followed by a final normalization, yielding the key (here $d_k=C$):
\begin{equation}
\mathbf{k}_t \;=\; \frac{ \tfrac{1}{2}\,\mathbf{m}_t + \tfrac{1}{2}\,\mathbf{x}_t }{ \left\lVert \tfrac{1}{2}\,\mathbf{m}_t + \tfrac{1}{2}\,\mathbf{x}_t \right\rVert_2 + \varepsilon } \in \mathbb{R}^{d_k}.
\end{equation}

At query time, we form a query vector $\mathbf{q}_t$ from the current frame using the same fusion, compute cosine similarities $s_i=\cos(\mathbf{q}_t,\mathbf{k}_i)$ to all keys, and select the top-$k$ indices $\mathcal{N}_k(\mathbf{q}_t)$. We then sample one key from this shortlist with a similarity-biased distribution:
\begin{equation}
p(i\mid \mathbf{q}_t)=\frac{\exp\!\left(s_i/\tau\right)}{\sum_{j\in \mathcal{N}_k(\mathbf{q}_t)} \exp\!\left(s_j/\tau\right)}\,,\qquad i\in \mathcal{N}_k(\mathbf{q}_t),
\end{equation}
where $\tau>0$ is a temperature. This ``retrieve-then-sample'' procedure preserves exploration among near-matches while favoring the most semantically similar experiences.

Figure~\ref{fig:efn-retrieval} illustrates the structure EFN exploits in the experience bank: instruction embeddings $\ell_\tau$ cluster by task type, making the instruction-filtered candidate set $\mathcal{R}_n$ small and focused, while retrieved keys form tight local neighborhoods around each query in the visual space.

\subsection{Learning with Residual Policy Optimization}
\label{sec:residual-policy}
\noindent \textbf{Problem setup.}
EFN learns to \emph{nudge} the base policy by recalling a relevant past experience and adjusting the current action so that the next observation resembles ``what happened next'' in that experience. At step $t$, the inputs are the current visual features $\mathbf{F}_t$ and the base policy's action $\mathbf{a}_t^{(0)}$, together with a retrieved experience step $(\hat{\mathbf{F}}, \hat{\mathbf{a}}, \hat{\mathbf{F}}^{+})$ and its rollout-level instruction embedding $\ell$ (retrieval is defined in the previous subsection). EFN's actor outputs a residual $\Delta\mathbf{a}_t$; the executed control is
\begin{equation}
\mathbf{a}_t \;=\; \mathbf{a}_t^{(0)} \;+\; \Delta\mathbf{a}_t .
\end{equation}
Intuitively, $\mathbf{a}_t^{(0)}$ preserves the base policy's competence, while $\Delta\mathbf{a}_t$ provides an experience-informed correction conditioned on the retrieved transition.

Figure~\ref{fig:efn-residual-actions} visualizes EFN's behavior in latent action space. We project both the base actions $\mathbf{a}_t^{(0)}$ and EFN-corrected actions $\mathbf{a}_t$ to 2D using PCA and draw arrows for the residuals $\Delta \mathbf{a}_t$. The arrows form short displacements around the base actions, indicating that EFN acts as a fine-grained correction layer on top of the frozen policy rather than as a separate controller that overrides it.

\noindent \textbf{Dense Semantic Match Reward.}
To quantify the notion of ``match the experience's next outcome,'' we compare the realized next observation $s_{t+1}$ with the experience's successor frame $\hat{s}^{+}$ at the \emph{semantic} level. Let $\mathbf{u}(\cdot)$ be the mean--max fusion described earlier, applied to vision features to produce a unit vector. After executing $\mathbf{a}_t$, the environment yields $s_{t+1}$ with vision features $\mathbf{F}_{t+1}$.
We define a dense similarity reward
\begin{equation}
r^{\text{sem}}_t \;=\; \cos\!\big(\mathbf{u}(\mathbf{F}_{t+1}),\, \mathbf{u}(\hat{\mathbf{F}}^{+})\big).
\end{equation}
In practice, we also regularize the residual magnitude to avoid destabilizing the base behavior:
\begin{equation}
r_t \;=\; \lambda_{\text{sem}}\, r^{\text{sem}}_t \;-\; \lambda_{\text{res}} \,\lVert \Delta\mathbf{a}_t \rVert_2^2 ,
\end{equation}
with $\lambda_{\text{sem}},\lambda_{\text{res}}>0$. This reward directly encodes our training signal without requiring supervised residual labels. Because $r_t$ is defined purely from feature similarity rather than a success/failure flag, EFN can be trained on both successful and failed rollouts as long as they contain meaningful state transitions. 

\begin{figure}[t]
 \centering
 \includegraphics[height=3.7cm,keepaspectratio]{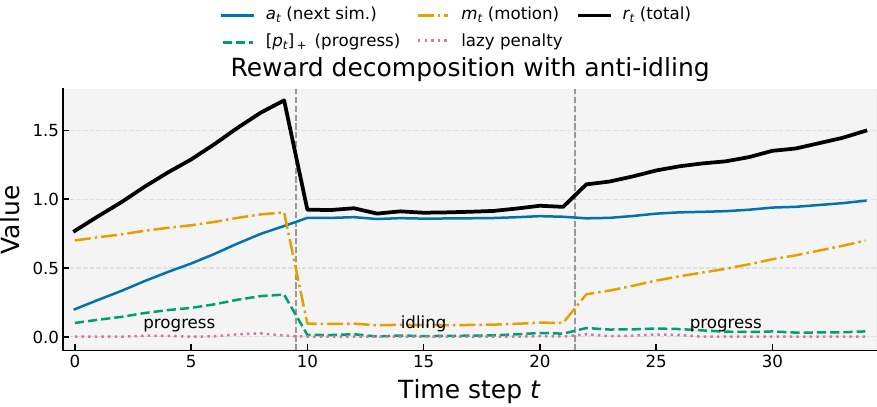}
 \caption{
Reward decomposition along a representative rollout under EFN's shaped objective.
Early in the trajectory, the agent makes progress and receives high reward;
when it idles near a good view, the lazy penalty suppresses $r_t$;
once it moves toward the retrieved successor again, the progress and motion terms dominate.
 }
 \label{fig:efn-reward-decomp}
 \vspace{-5pt}
\end{figure}

\noindent \textbf{SAC objective.}
We train EFN with Soft Actor--Critic, conditioning both actor and critics on the current and retrieved experience context. Let
\begin{equation}
\mathbf{c}_t \;=\; \mathrm{enc}\!\big(\mathbf{F}_t,\, \mathbf{a}_t^{(0)},\, \hat{\mathbf{F}},\, \hat{\mathbf{a}},\, \ell\big)
\end{equation}
be a learned context representation (the base policy is frozen). The stochastic residual policy is $\pi_\phi(\Delta\mathbf{a}_t\,|\,\mathbf{c}_t)$, and the Q-functions $Q_{\theta_1},Q_{\theta_2}$ evaluate the corrected action $\mathbf{a}_t=\mathbf{a}_t^{(0)}+\Delta\mathbf{a}_t$ under $\mathbf{c}_t$. The critic targets use the soft Bellman backup with target networks $\bar{\theta}_i$:
\begin{equation}
\begin{aligned}
y_t &= r_t 
 + \gamma\,\mathbb{E}_{\Delta\mathbf{a}_{t+1}\!\sim\!\pi_\phi(\cdot|\mathbf{c}_{t+1})}\!\Big[ \\
& \min_{i} Q_{\bar{\theta}_i}\!(\mathbf{c}_{t+1},\,\mathbf{a}_{t+1}^{(0)}+\Delta\mathbf{a}_{t+1})
 - \alpha \log \pi_\phi(\Delta\mathbf{a}_{t+1}|\mathbf{c}_{t+1}) \Big].
\end{aligned}
\end{equation}

The critic loss is the standard squared error:
\begin{equation}
\mathcal{L}_{\text{critic}}(\theta_1,\theta_2) \;=\; 
\sum_{i=1,2} \,\mathbb{E}\Big[\big(Q_{\theta_i}(\mathbf{c}_t,\, \mathbf{a}_t^{(0)}+\Delta\mathbf{a}_t) - y_t\big)^2\Big].
\end{equation}
The actor minimizes the entropy-regularized objective
\begin{equation}
\begin{aligned}
\mathcal{L}_{\text{actor}}(\phi)
=\;& 
\mathbb{E}_{\Delta\mathbf{a}_t \sim \pi_\phi(\cdot\,|\,\mathbf{c}_t)}\!
\Big[\, \alpha \,\log \pi_\phi(\Delta\mathbf{a}_t\,|\,\mathbf{c}_t) \;-\; \\[2pt]
& \min_{i=1,2} Q_{\theta_i}\!\big(\mathbf{c}_t,\, \mathbf{a}_t^{(0)}+\Delta\mathbf{a}_t\big) \Big].
\end{aligned}
\end{equation}
with temperature $\alpha$ optionally tuned to maintain a target entropy. During training, gradients do not flow through the retrieval targets $\hat{\mathbf{F}}$ and $\hat{\mathbf{F}}^{+}$; updates are confined to EFN's actor, critics, and the context encoder.

\begin{figure}[t]
 \centering
 \includegraphics[height=3.6cm,keepaspectratio]{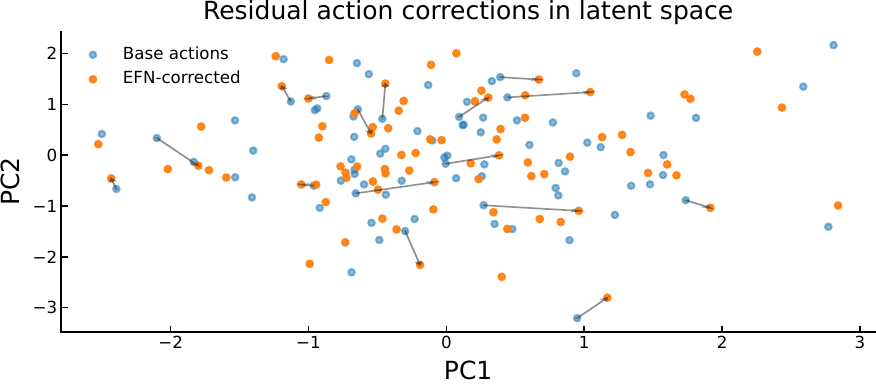}
 \caption{
 Residual corrections in latent action space.
 We project base actions $\mathbf{a}_t^{(0)}$ and EFN-corrected actions
 $\mathbf{a}_t = \mathbf{a}_t^{(0)} + \Delta \mathbf{a}_t$ to 2D using PCA.
 Each arrow connects a base action (yellow) to its corrected counterpart (blue),
 showing that EFN applies small, local residuals $\Delta \mathbf{a}_t$
 rather than replacing the base policy outright.
 }
 \label{fig:efn-residual-actions}
  \vspace{-5pt}
\end{figure}

\noindent \textbf{Reward shaping with anti-idling penalty.}
To retain absolute credit for matching the retrieved next observation while penalizing ``similar but no progress'' behavior and encouraging shorter rollouts, we define semantic similarity ($\mathrm{sim}\in[0,1]$)
\begin{equation}
\begin{aligned}
s_t^{\mathrm{next}} &= \mathrm{sim}(\mathbf{F}_{t+1},\,\hat{\mathbf{F}}^{+}),\\
s_t^{\mathrm{cur}} &= \mathrm{sim}(\mathbf{F}_{t},\,\hat{\mathbf{F}}),\\
s_t^{\mathrm{stay}} &= \mathrm{sim}(\mathbf{F}_{t+1},\,\mathbf{F}_{t}).
\end{aligned}
\end{equation}
We introduce auxiliary terms (with $[x]_+=\max(x,0)$ and tolerance $\varepsilon>0$):
\begin{equation}
a_t=s_t^{\mathrm{next}},
p_t=s_t^{\mathrm{next}}-s_t^{\mathrm{cur}},
m_t=1-s_t^{\mathrm{stay}},
n_t=[\,\varepsilon - p_t\,]_+.
\end{equation}
The dense reward becomes
\begin{equation}
\label{eq:efn-reward}
\begin{aligned}
r_t =\;&
w_{\text{abs}}\, a_t
+ w_{\text{prog}}\, [p_t]_+
+ w_{\text{mot}}\, m_t \\
& -\, w_{\text{lazy}}\,\big(s_t^{\mathrm{next}}\, n_t\, s_t^{\mathrm{stay}}\big)
- \lambda_{\text{time}}.
\end{aligned}
\end{equation}
where $w_{\text{abs}},w_{\text{prog}},w_{\text{mot}},w_{\text{lazy}}\!\ge\!0$ and $\lambda_{\text{time}}\!\ge\!0$ are scalars.
Equation~\eqref{eq:efn-reward} preserves an \emph{absolute similarity} bonus ($a_t$), adds a \emph{progress} term ($[p_t]_+$) to credit genuine improvement toward the retrieved next state, encourages \emph{non-trivial motion} ($m_t$) to avoid degenerate idling, and introduces an \emph{anti-idling penalty} that activates only when the prediction is already similar yet shows negligible improvement and little change between consecutive frames. The per-step cost $\lambda_{\text{time}}$ favors shorter successful trajectories. Conceptually, EFN therefore learns a residual policy that steers $\mathbf{F}_{t+1}$ toward $\hat{\mathbf{F}}^{+}$.

\subsection{Deployment Retrieval and Experience Growth}
\label{sec:deployment}
\noindent \textbf{Goal and key differences.}
At deployment, EFN recalls and reuses prior experiences while \emph{not} updating either the base policy weights or its own. The inference pipeline mirrors training but differs in three ways. First, retrieval is \emph{task-filtered}: we restrict matches to rollouts whose instruction embeddings are close to the current task. Second, we \emph{prioritize efficient rollouts}: shorter trajectories receive higher selection priority because they typically contain fewer redundant actions and lead to faster completion. Third, we \emph{grow the bank online}: after a rollout finishes, its steps are inserted into the bank so that future episodes can recall them. Post-deployment adaptation therefore comes purely from memory growth, not gradient updates.

\noindent \textbf{Instruction-filtered candidate set.}
Given a task description, we compute an instruction embedding with the VLA's language encoder, denoted $\ell^\star$. We compare $\ell^\star$ with all stored rollout-level embeddings $\{\ell_{\tau_j}\}$ using cosine similarity and select the top-$n$ rollouts:
\begin{equation}
\mathcal{R}_n \;=\; \mathrm{Top}\text{-}n\;\big\{\cos(\ell^\star,\ell_{\tau_j})\big\}_j .
\end{equation}
All step-level entries from these rollouts form the \emph{candidate experience set} $\mathcal{C}$, which is the only pool from which we retrieve during this episode.

\noindent \textbf{Step-wise retrieval with efficiency prior.}
At step $t$, we form a visual query $\mathbf{q}_t=\mathbf{u}(\mathbf{F}_t)$ using the mean--max fusion $\mathbf{u}(\cdot)$ defined previously. Each candidate $i\in\mathcal{C}$ has a key $\mathbf{k}_i$, a successor feature $\hat{\mathbf{F}}^{+}_i$, and belongs to a rollout $\rho(i)$ with total length $L_{\rho(i)}$. We score candidates by combining semantic similarity with an efficiency prior that favors shorter rollouts. Let $s_i=\cos(\mathbf{q}_t,\mathbf{k}_i)$ and define a normalized length prior $g(L_{\rho(i)})\in[0,1]$ that decreases with $L_{\rho(i)}$ (e.g., $g(L)=\exp[-\beta\,L/\bar{L}]$ with temperature $\beta$ and reference length $\bar{L}$). The combined score is
\begin{equation}
\tilde{s}_i \;=\; \lambda\, s_i \;+\; (1-\lambda)\, g\!\big(L_{\rho(i)}\big), \qquad \lambda\in[0,1].
\end{equation}
We take the top-$k$ candidates by $\tilde{s}_i$ and sample one with softmax:
\begin{equation}
p(i\mid t)\;=\;\frac{\exp\!\left(\tilde{s}_i/\tau\right)}{\sum_{j\in\mathcal{N}_k(t)}\exp\!\left(\tilde{s}_j/\tau\right)}\,,\qquad i\in\mathcal{N}_k(t),
\end{equation}
where $\tau$ is a temperature and $\mathcal{N}_k(t)$ denotes the $k$ highest-scoring items at step $t$. This procedure preserves exploration among near-matches while preferring memories that both look similar and come from efficient behaviors.

\noindent \textbf{Action correction and execution.}
Conditioned on the current context and the sampled experience $(\hat{\mathbf{F}}_i,\hat{\mathbf{a}}_i,\hat{\mathbf{F}}^{+}_i)$, EFN predicts a residual $\Delta\mathbf{a}_t$ and executes $\mathbf{a}_t=\mathbf{a}_t^{(0)}+\Delta\mathbf{a}_t$ as in training. At inference, all critics and the residual policy remain fixed; in practice, we use the deterministic mean action (or a low-temperature sample) to reduce variance. Conceptually, the correction is chosen to steer the next observation toward the stored successor $\hat{\mathbf{F}}^{+}_i$, reusing successful transitions without further learning updates.

\noindent \textbf{Online experience growth.}
After each episode, we insert the resulting rollout into the experience bank: the episode-level instruction embedding $\ell^\star$ is stored together with all step tuples $\big(\mathbf{F}_t, \mathbf{k}_t, \mathbf{a}_t^{(0)}\big)$ computed under the same mean--max keying scheme. At deployment time, however, we append only rollouts that successfully reach the goal, unlike the training phase, which may also admit near-successful or failed trajectories (Appendix Section\ref{sec:success-failure}). These successful episodes serve as high-quality references for future retrieval. When operating under an experience budget, standard retention strategies, such as reservoir sampling or recency-aware replacement, can be applied without modifying the retrieval or learning mechanisms described above.

\begin{table*}[ht]
\caption{Performance on the LIBERO benchmark, averaged over three seeds.}
\centering
\setlength{\abovecaptionskip}{-3pt}
\setlength{\belowcaptionskip}{0pt}
\small
\renewcommand{\arraystretch}{0.85}
\begin{adjustbox}{width=\linewidth}
\begin{tabular}{l cc cc cc cc cc}
\toprule
\multirow{2}{*}{Method} &
\multicolumn{2}{c}{Spatial} &
\multicolumn{2}{c}{Object} &
\multicolumn{2}{c}{Goal} &
\multicolumn{2}{c}{Long} &
\multicolumn{2}{c}{Average} \\
\cmidrule(lr){2-3}\cmidrule(lr){4-5}\cmidrule(lr){6-7}\cmidrule(lr){8-9}\cmidrule(lr){10-11}
& Succ.\,$\uparrow$ & Step\,$\downarrow$
& Succ.\,$\uparrow$ & Step\,$\downarrow$
& Succ.\,$\uparrow$ & Step\,$\downarrow$
& Succ.\,$\uparrow$ & Step\,$\downarrow$
& Succ.\,$\uparrow$ & Step\,$\downarrow$ \\
\midrule
\multicolumn{11}{c}{\textbf{Base: OpenVLA}} \\
\midrule
OpenVLA~\cite{kim2024openvla}         & 84.7 & 119.5 & 88.4 & 163.7 & 79.2 & 121.5 & 53.7 & 275.9 & 76.5 & 160.2 \\
+kNN-RAG         & 82.3 & 125.1 & 85.6 & 171.2 & 74.8 & 128.3 & 50.1 & 282.7 & 73.2 & 166.4 \\
+ResAct~\cite{liu2025visual}         & 86.1 & 117.3 & 89.7 & 160.5 & 82.5 & 119.7 & 58.4 & 271.5 & 79.2 & 158.6 \\
+R2A~\cite{goyal2022retrieval}           & 87.5 & 114.8 & 91.1 & 157.2 & 84.6 & 116.9 & 63.2 & 265.3 & 81.6 & 156.3 \\
+GC-TTT~\cite{sun2020test}         & 85.2 & 121.9 & 87.9 & 165.8 & 80.7 & 123.6 & 55.9 & 277.1 & 77.4 & 162.8 \\
\textbf{+EFN (Vol=300)} & 88.5 & 115.4 & 91.3 & 158.8 & 85.7 & 117.0 & 72.1 & 267.2 & 84.4 & 160.0 \\
\textbf{+EFN (Vol=1000)} & \textbf{89.9} & \textbf{109.0} & \textbf{92.2} & \textbf{156.1} & \textbf{89.2} & \textbf{115.2} & \textbf{76.5} & \textbf{261.9} & \textbf{87.0} & \textbf{156.7} \\
\midrule
\multicolumn{11}{c}{\textbf{Base: UniVLA}} \\
\midrule
UniVLA~\cite{bu2025univla}          & 96.5 & 112.7 & 96.8 & 159.0 & 95.6 & 124.9 & 92.0 & 264.5 & 95.2 & 164.2 \\
+kNN-RAG         & 94.2 & 118.3 & 95.7 & 165.1 & 93.1 & 128.6 & 87.5 & 270.3 & 92.6 & 168.9 \\
+ResAct~\cite{liu2025visual}         & 97.1 & 109.6 & 97.4 & 154.7 & 96.2 & 122.1 & 89.8 & 260.7 & 95.1 & 160.0 \\
+R2A~\cite{goyal2022retrieval}           & 97.5 & 107.2 & 97.8 & 152.3 & 96.6 & 120.5 & 91.3 & 257.9 & 95.8 & 158.0 \\
+GC-TTT~\cite{sun2020test}         & 96.8 & 110.1 & 97.1 & 156.8 & 95.9 & 123.4 & 89.1 & 266.1 & 94.7 & 162.1 \\
\textbf{+EFN (Vol=300)} & 97.7 & 103.4 & 97.9 & 151.4 & 97.2 & 120.1 & 93.7 & 253.8 & 96.6 & 156.2 \\
\textbf{+EFN (Vol=1000)} & \textbf{98.2} & \textbf{102.1} & \textbf{98.2} & \textbf{145.8} & \textbf{97.6} & \textbf{117.8} & \textbf{94.6} & \textbf{242.5} & \textbf{97.2} & \textbf{151.3} \\
\midrule
\multicolumn{11}{c}{\textbf{Base: GO-1}} \\
\midrule
GO-1~\cite{bu2025agibot}           & 96.3 & 111.6 & 97.4 & 160.2 & 95.6 & 118.9 & 89.3 & 269.1 & 94.7 & 163.1 \\
+kNN-RAG         & 94.1 & 116.8 & 95.9 & 167.5 & 93.4 & 122.7 & 85.1 & 274.9 & 92.1 & 168.0 \\
+ResAct~\cite{liu2025visual}         & 96.8 & 108.9 & 97.6 & 156.1 & 96.1 & 119.8 & 88.7 & 264.7 & 94.8 & 160.3 \\
+R2A~\cite{goyal2022retrieval}           & 97.3 & 106.5 & 97.9 & 153.6 & 96.5 & 118.2 & 90.2 & 261.5 & 95.5 & 158.1 \\
+GC-TTT~\cite{sun2020test}         & 96.6 & 113.2 & 97.2 & 159.5 & 95.8 & 121.1 & 88.9 & 267.3 & 94.6 & 163.3 \\
\textbf{+EFN (Vol=300)} & 97.5 & 107.8 & 98.1 & 154.5 & 96.8 & 115.6 & 91.2 & 258.5 & 95.9 & 157.5 \\
\textbf{+EFN (Vol=1000)} & \textbf{98.1} & \textbf{105.3} & \textbf{98.5} & \textbf{151.7} & \textbf{97.3} & \textbf{114.1} & \textbf{92.8} & \textbf{253.2} & \textbf{96.7} & \textbf{154.8} \\
\bottomrule
\label{libero}
\end{tabular}
\end{adjustbox}
\vspace{-5mm}
\end{table*}

\begin{figure*}[t]
   \centering
   \setlength{\abovecaptionskip}{2pt}
\setlength{\belowcaptionskip}{-4pt}
   \includegraphics[width=1\linewidth]{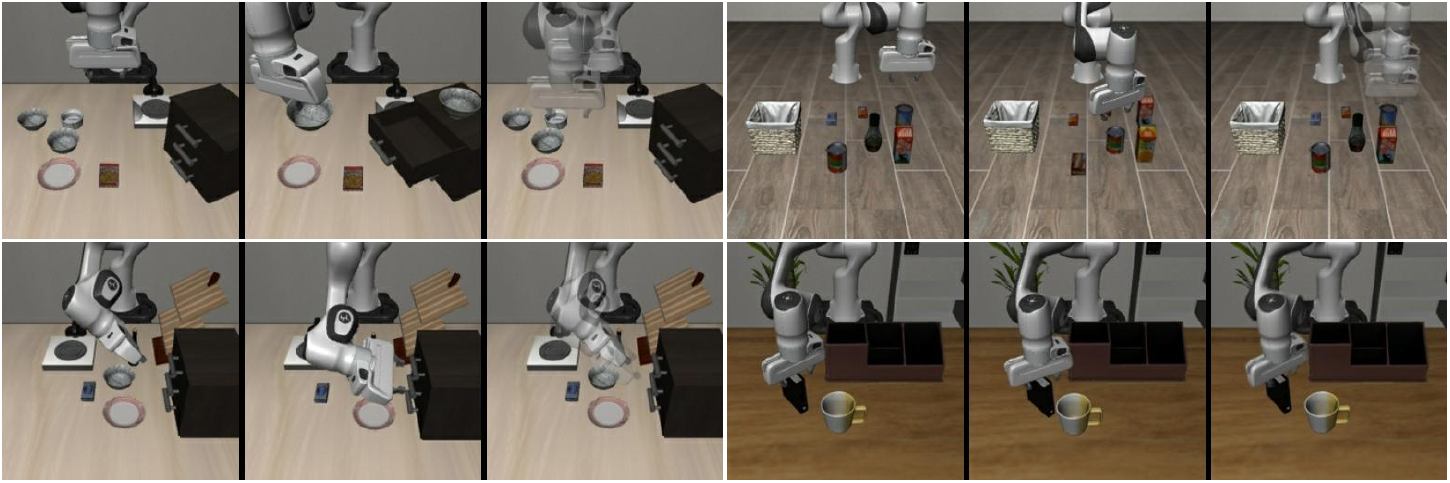}
 \caption{Visualization of EFN's residual corrections.
Left: current observation. Middle: retrieved frame. Right: corrected (translucent) versus base action.
Even when the retrieved scene differs from the current one (top row), EFN can still extract actionable guidance.
}
 \label{fig:vis}
 
\end{figure*}

\begin{table*}[ht]
\caption{Real-world performance on the AgiBot-G1 platform with the GO-1 policy~\cite{bu2025agibot}.}
\label{agibot}
\centering
\setlength{\abovecaptionskip}{-3pt}
\setlength{\belowcaptionskip}{0pt}
\small
\renewcommand{\arraystretch}{0.85}
\begin{adjustbox}{width=\linewidth}
\begin{tabular}{l cc cc cc cc cc}
\toprule
\multirow{2}{*}{Method} &
\multicolumn{2}{c}{BottlePlace} &
\multicolumn{2}{c}{ShelfSort} &
\multicolumn{2}{c}{StockLift} &
\multicolumn{2}{c}{DrawerStore} &
\multicolumn{2}{c}{Average} \\
\cmidrule(lr){2-3}\cmidrule(lr){4-5}\cmidrule(lr){6-7}\cmidrule(lr){8-9}\cmidrule(lr){10-11}
& Succ.\,$\uparrow$ & Step\,$\downarrow$
& Succ.\,$\uparrow$ & Step\,$\downarrow$
& Succ.\,$\uparrow$ & Step\,$\downarrow$
& Succ.\,$\uparrow$ & Step\,$\downarrow$
& Succ.\,$\uparrow$ & Step\,$\downarrow$ \\
\midrule
GO-1~\cite{bu2025agibot}           & 47.3 & 407.0 & 34.0 & 442.1 & 16.0 & 389.2 & 5.3 & 728.9 & 25.8 & 491.8 \\
+kNN-RAG         & 35.3 & 435.2 & 25.3 & 468.7 & 12.0 & 405.1 & 0.0 & 752.3 & 18.2 & 515.3 \\
+ResAct~\cite{liu2025visual}          & 52.7 & 418.9 & 41.3 & 431.4 & 22.7 & 396.8 & 8.0 & 715.6 & 31.2 & 490.7 \\
+R2A~\cite{goyal2022retrieval}           & 61.3 & 399.1 & 51.3 & 415.7 & 31.3 & 372.5 & 14.0 & 689.2 & 39.5 & 469.1 \\
+GC-TTT~\cite{sun2020test}          & 49.3 & 412.5 & 36.7 & 448.9 & 18.7 & 393.7 & 0.0 & 735.1 & 26.2 & 497.6 \\
\textbf{+EFN (Vol=300)}  & 69.3 & 382.6 & 54.7 & 410.5 & 42.0 & 355.1 & 37.3 & 668.3 & 50.8 & 454.1 \\
\textbf{+EFN (Vol=1000)} &
\textbf{82.0} & \textbf{364.2} &
\textbf{74.7} & \textbf{392.9} &
\textbf{65.3} & \textbf{339.9} &
\textbf{58.7} & \textbf{643.4} &
\textbf{70.2} & \textbf{435.1} \\
\bottomrule
\end{tabular}
\end{adjustbox}
\end{table*}

\vspace{-5pt}
\section{Experiments}
\vspace{-2pt}
\subsection{Experimental setup.}
\vspace{-2pt}
We evaluate EFN on the \texttt{LIBERO} benchmark~\cite{liu2023libero} with three pretrained VLAs: OpenVLA~\cite{kim2024openvla}, UniVLA~\cite{bu2025univla}, and GO-1~\cite{bu2025agibot}. All models use the same visual preprocessing (center crop followed by resize to $256$). UniVLA exposes $256$ visual tokens and $4$ latent action tokens per step; EFN takes the current tokens and retrieved tokens, predicts a residual in latent space, and adds it to the base latent action, which is then decoded by the \emph{frozen} action head. Training details are provided in Appendix Section\ref{details}.

\noindent \textbf{Evaluation protocol.}
At test time, we make the policy deterministic by applying $\tanh$ to the actor mean. We report success rate and, conditional on success, the average number of steps (lower is better) with horizon $H{=}320$. The experience bank used at evaluation mirrors training and is queried via cosine nearest neighbors over pooled visual embeddings. After each successful rollout, we append the episode to the bank and prioritize shorter successful episodes; no gradient updates are performed during evaluation.

\noindent \textbf{Baselines and variants.}
We compare each backbone with four baselines and our EFN variants, chosen to instantiate the three paradigms from Sec.~\ref{sec:related}. \textit{kNN-RAG} is a non-parametric retrieval-only controller that retrieves the nearest past step from the shared bank and directly copies its action. \textit{ResAct}~\cite{liu2025visual} is a residual-only baseline that learns an additive correction on top of the frozen VLA action but does not use retrieval or similarity-shaped rewards. \textit{R2A}~\cite{goyal2022retrieval} adapts retrieval-augmented RL to VLAs: the policy is conditioned on retrieved trajectories and optimized end-to-end, updating the backbone. \textit{GC-TTT}~\cite{sun2020test} is a goal-conditioned test-time training baseline that updates the backbone online using retrieved rollouts. Additional baseline details are provided in Appendix Section\ref{sec:sim-setup}.

All retrieval-based methods share the \emph{same} experience bank and retrieval interface, differing only in how retrieval is used (direct action copying, residual correction, or backbone updates). EFN is evaluated with two bank capacities, \texttt{Vol}\,$\in\{300,1000\}$, denoting the maximum number of stored transitions \emph{per task}. Real-world experiments run on the AgiBot-G1 platform with GO-1~\cite{bu2025agibot}. Runtime and memory overheads, reported in Appendix Section\ref{cost}, remain small relative to the frozen VLA forward pass. Each setting uses 50 episodes across 3 seeds, and we report the average.

\vspace{-3pt}
\subsection{Benchmark and Analysis}
\vspace{-2pt}
\noindent \textbf{Results on LIBERO.}
Table~\ref{libero} summarizes the results on \texttt{LIBERO}. Across all three backbones, EFN consistently improves both success and efficiency over the frozen VLAs. The retrieval-only baseline kNN-RAG often degrades performance, suggesting that copying actions from nearest neighbors is brittle. ResAct provides modest gains, indicating that residual refinement helps but remains limited without episodic context. R2A benefits from retrieval-augmented RL but requires updating the large backbone and still underperforms EFN under the same interaction budget. GC-TTT yields small or negative improvements, reflecting the instability of aggressive test-time finetuning on non-i.i.d.\ rollouts. EFN with a small bank (\texttt{Vol=300}) already matches or surpasses all baselines, while \texttt{Vol=1000} offers further, though diminishing, gains. Figure~\ref{fig:vis} illustrates how EFN's residual actions correct the base policy's outputs.

\noindent \textbf{Real-world results.}
Table~\ref{agibot} reports results on AgiBot-G1 with GO-1 on four real-world tasks: BottlePlace, ShelfSort, StockLift, and DrawerStore, ordered roughly from easy to hard. Detailed settings for these tasks are provided in Appendix Section\ref{sec:realworld-setup}. The trends mirror simulation: kNN-RAG consistently hurts performance; ResAct and R2A offer moderate gains; and GC-TTT is unstable and fails on the hardest task. EFN achieves the highest average success with fewer steps, and the gap between \texttt{Vol=300} and \texttt{Vol=1000} is smaller than in simulation, suggesting that a moderate bank already covers the main task variations. All EFN results are obtained with a frozen GO-1 backbone; adaptation arises solely from the experience bank and the residual head, whereas R2A and GC-TTT rely on gradient-based updates. We include visualizations of the real-world experiments in Appendix Section\ref{vis}.

\subsection{Ablation Study}
We further evaluate four controlled variants on LIBERO (Figure~\ref{fig:abla}); full ablation numbers are provided in Appendix Table~\ref{sec:success-failure}.

\noindent \textbf{Effect of SAC.} Removing SAC (\emph{w/o SAC}) and replacing it with a value-only critic harms both success and efficiency, supporting the use of entropy-regularized optimization.

\noindent \textbf{Effect of dense rewards.} Dropping similarity-shaped dense rewards (\emph{w/o dense}) and relying only on sparse task returns slows learning and reduces final performance.

\begin{figure}[t]
 \centering
 \setlength{\abovecaptionskip}{-1pt}
\setlength{\belowcaptionskip}{-3pt}
 \includegraphics[width=1\linewidth]{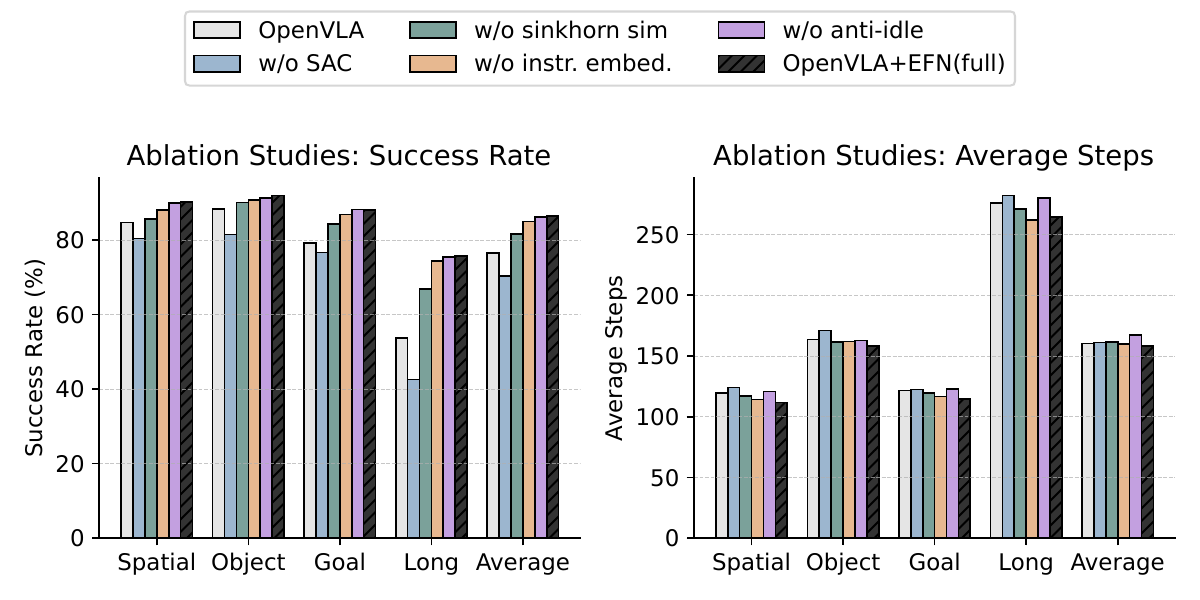}
 \caption{Ablation studies of EFN on the LIBERO benchmark.}
 \label{fig:abla}
 \vspace{-5pt}
\end{figure}

\begin{table}[t]
 \centering
 \small
 \setlength{\tabcolsep}{5pt}
 \caption{Inference-time efficiency comparison for real-world deployment.
Change is computed relative to the base GO-1.}
 \label{tab:efn-latency}
 \begin{tabular}{lcc}
  \toprule
  Metric & GO-1 & GO-1 + EFN (ours) \\
  \midrule
  Per-step latency (ms) $\downarrow$
   & \textbf{35.7}
   & 37.2 {\color{red}{(+4.2\%)}} \\

  Avg. steps / episode $\downarrow$
   & 491.8
   & \textbf{435.1} {\color{ForestGreen}{(-11.5\%)}} \\

  Episode time (s) $\downarrow$
   & 17.6
   & \textbf{16.2} {\color{ForestGreen}{(-7.9\%)}} \\
  \bottomrule
 \end{tabular}
 \vspace{-5pt}
\end{table}

\noindent \textbf{Effect of instruction.} Disabling instruction-level filtering (\emph{w/o instr}) degrades performance, indicating that instruction-aware retrieval is important.

\noindent \textbf{Effect of anti-idle.} Removing the anti-idling term (\emph{w/o anti-idle}) reduces performance, indicating that idleness penalties matter.

\noindent \textbf{Effect of residual action.} As further analyzed in Appendix Section\ref{app:direct-action}, replacing the residual correction with direct action prediction degrades performance, which we attribute to discarding the base policy's generalization and overburdening the lightweight actor with modeling the full action space.

\noindent \textbf{Inference time.} EFN improves performance while remaining efficient. Although each step introduces a modest $4.2\%$ latency overhead, EFN reduces the average episode length by $11.5\%$, resulting in an overall $7.9\%$ reduction in total episode time. In other words, EFN achieves higher success rates while also reducing end-to-end episode time.

\vspace{-5pt}
\section{Conclusion}
\vspace{-2pt}
Inspired by the human experience of déjà vu, we introduce the Experience Feedback Network, a retrieval-conditioned residual module that augments a frozen vision--language--action policy with an experience bank. EFN retrieves task-relevant transitions in a joint vision--language space and trains a lightweight residual controller with similarity-shaped reinforcement learning, enabling post-deployment adaptation through memory updates instead of backbone finetuning. On LIBERO and a real-world AgiBot-G1 platform, EFN consistently outperforms retrieval-only, residual-only, retrieval-augmented RL, and test-time training baselines under the same interaction budget.

\section*{Acknowledgments}
This research is supported by the Fundamental Research Funds for the Central Universities (project number YG2024ZD06), NSFC (No. 62176155), and Shanghai Municipal Science and Technology Major Project (2021SHZDZX0102).

{
  \small
  \bibliographystyle{ieeenat_fullname}
  \bibliography{main}
}

\appendix
\counterwithin{figure}{section}
\counterwithin{table}{section}
\counterwithin{equation}{section}
\counterwithin{algorithm}{section}
\renewcommand{\thealgorithm}{\thesection.\arabic{algorithm}}

\section{Overview}
This appendix is organized as follows. Section~\ref{details} describes the detailed settings of our method, baselines, and experiments. Section~\ref{rationale} elaborates on the rationale, motivation, and intuition behind EFN. Section~\ref{dis} discusses the method and its broader implications. Section~\ref{exp} presents supplementary experimental results and analysis. Section~\ref{vis} provides additional visualizations of EFN in action.

\section{Details of Method and Experiment}
\label{details}
This section provides implementation and experimental details; the rationale behind these design choices is discussed in Section~\ref{rationale}.
\subsection{Experience Bank Implementation Details}
\label{app:experience-bank-impl}

\paragraph{Step-level representation.}
Recall that EFN operates on step-level transitions from rollouts
$\tau=(s_1,a_1,\dots,s_T,a_T)$ and augments a frozen VLA policy using an experience bank.
Conceptually, each stored step $E_{m,n}$ from rollout $m$ at time $n$ contains
\begin{equation}
E_{m,n} = \big(\ell_{\tau_m},\,\mathbf{F}_{m,n},\,\mathbf{k}_{m,n},\,\mathbf{a}^{(0)}_{m,n}\big),
\end{equation}
where $\ell_{\tau_m}$ is the instruction embedding for rollout $\tau_m$, $\mathbf{F}_{m,n}\in\mathbb{R}^{L\times C}$ is the VLA vision encoder feature map for frame $s_{m,n}$, $\mathbf{k}_{m,n}\in\mathbb{R}^{d_k}$ is the mean--max fused retrieval key defined in main-paper Sec.~3.2, and $\mathbf{a}^{(0)}_{m,n}$ is the raw action produced by the frozen base policy at that step.

In our implementation, the bank is stored as a line-based JSON file together with accompanying feature files, where each JSON line encodes one step and includes a rollout identifier, a step index, and paths to the visual and action features.
The corresponding visual feature file stores the precomputed retrieval key
$\mathbf{k}_{m,n}$ as a $d_k$-dimensional vector (e.g., \verb|visual_embed_meanmax|
with $d_k=4096$), together with any additional visual features that may be needed.
The action file stores the base policy's latent action representation
(e.g., a $4\times d_a$ tensor of action tokens) and the associated discrete token IDs
(\verb|generated_ids|) from which the continuous control command
$\mathbf{a}^{(0)}_{m,n}$ is decoded.

\begin{figure}[h]
  \centering
  \includegraphics[width=\linewidth]{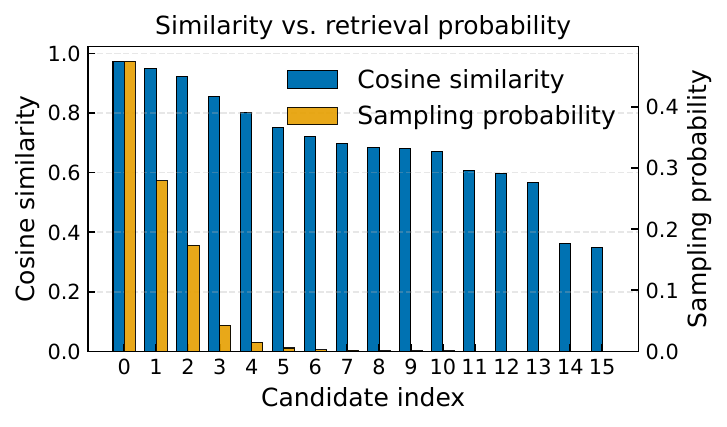}
  \caption{
    Cosine similarities (left bars) and corresponding
    softmax sampling probabilities (right bars) over the top candidates
    in the experience bank. Higher similarity induces higher
    sampling probability while still preserving exploration.
  }
  \label{fig:sim-prob}
\end{figure}

\paragraph{Rollout-wise organization and indexing.}
Experiences are inserted into the bank at every non-blank step where the robot executes a control action. We explicitly omit the initial waiting steps during simulator loading and robot warm-up from both training and evaluation; these steps are never written to the JSON file and therefore never appear as candidates in retrieval.

The JSON file implicitly groups steps by \verb|rollout_id| and \verb|step_idx|.
At load time, we construct two data structures:
(i) a mapping
\begin{equation}
\texttt{index} : (\texttt{rollout\_id},\,\texttt{step\_idx}) \mapsto E_{m,n},
\end{equation}
implemented as a hash table from the pair \((m,n)\) to the corresponding metadata record,
and (ii) a dense key matrix
\begin{equation}
\mathbf{K} \in \mathbb{R}^{N \times d_k},
\end{equation}
obtained by stacking all keys $\mathbf{k}_{m,n}$ for the $N$ stored steps.
This enables constant-time lookup of a particular step and efficient vectorized similarity computation against all keys.

For convenience, we also expose a ``successor'' operator $E_{m,n+1}$ that returns the
immediate next step of a given transition whenever it exists:
\begin{equation}
E_{m,n+1} = \texttt{index}\big(\texttt{rollout\_id}=m,\;\texttt{step\_idx}=n+1\big).
\end{equation}
We use $E_{m,n+1}$ as the semantic target successor frame when shaping the progress reward:
the retrieved step provides a reference for ``what should happen next'' if the agent
were to follow a previously successful trajectory.

\paragraph{Retrieval over the bank.}
Given a current observation $s_t$, we compute the vision encoder feature map
$\mathbf{F}_t$ and derive a query key $\mathbf{q}_t$ via the same mean--max fusion
and $\ell_2$-normalization used for $\mathbf{k}_{m,n}$ (main-paper Sec.~3.2).
We then compute cosine similarities between $\mathbf{q}_t$ and all stored keys:
\begin{equation}
s_i = \cos(\mathbf{q}_t, \mathbf{k}_i), \quad i \in \{1,\dots,N\}.
\end{equation}


In the probabilistic variant used for EFN training, we follow the top-$k$ sampling scheme described in the main text: we first select a shortlist
$\mathcal{N}_k(\mathbf{q}_t)$ of the $k$ highest-similarity keys and sample an index
using a temperature-scaled softmax,
\begin{equation}
p(i\mid \mathbf{q}_t) =
\frac{\exp\big(s_i/\tau\big)}
     {\sum_{j \in \mathcal{N}_k(\mathbf{q}_t)} \exp\big(s_j/\tau\big)},\quad
i \in \mathcal{N}_k(\mathbf{q}_t),
\end{equation}
with a fixed temperature $\tau>0$ (we use $\tau=0.05$ in all experiments).
As shown in Figure~\ref{fig:sim-prob}, this ``retrieve-then-sample'' design encourages exploration among near-matches while
still biasing toward the most semantically similar experiences.

\paragraph{Usage during training and deployment.}
During offline pretraining of EFN, the experience bank is constructed from a mixture of successful, near-successful, and occasionally failed rollouts, as discussed in Appendix Section~\ref{dis}. At each environment step, EFN queries the bank, obtains a reference step $E_{m,n}$ and its successor $E_{m,n+1}$, and uses the successor observation as the semantic target for the similarity-shaped progress reward.
At deployment, the bank continues to grow: we append only the trajectories that reach
the goal, thereby gradually refining the set of available references without
ever updating the frozen VLA backbone weights.

Even failed or near-success rollouts can still contribute meaningful similarity structure,
since the reward is defined on feature transitions rather than binary outcome labels
(see Sec.~\ref{sec:success-failure} for analysis).

\subsection{Reward Shaping and SAC Training Details}
\label{app:reward-sac}

This subsection details the implementation of the similarity terms in main-paper Eq.~(13) and the Soft Actor--Critic (SAC)~\cite{haarnoja2018soft} training procedure used for EFN.

\paragraph{Similarity function \texorpdfstring{$\mathrm{sim}(\cdot,\cdot)$}{sim(·,·)}.}
For reward shaping, we instantiate $\mathrm{sim}(\cdot,\cdot)$ as a cosine-based similarity on the per-patch vision tokens.
Given two feature maps $X,Y\in\mathbb{R}^{256\times 4096}$ from the frozen VLA backbone, we first $\ell_2$-normalize each token along the channel dimension and form the cosine matrix
\begin{multline}
S = \mathrm{token\_cosine\_matrix}(X,Y) \\
= \mathrm{norm}(X)\,\mathrm{norm}(Y)^\top \in [-1,1]^{256\times 256}.
\end{multline}
We then compute a Sinkhorn-regularized matching~\cite{cuturi2013sinkhorn} between the two token sets with a Gibbs kernel $K=\exp(S/\varepsilon_{\text{OT}})$, uniform row and column marginals, and $n_{\text{iter}}$ Sinkhorn iterations.
The resulting transport plan $P\in\mathbb{R}^{256\times 256}$ satisfies approximately uniform marginals and we define the scalar score
\begin{equation}
\mathrm{score}(X,Y) \;=\; \sum_{i,j} P_{ij}\,S_{ij},
\end{equation}
followed by a linear rescaling to $[0,1]$:
\begin{equation}
\mathrm{sim}(X,Y) \;=\; \tfrac{1}{2}\,\big(\mathrm{clip}(\mathrm{score}(X,Y),-1,1)+1\big).
\end{equation}
In all experiments, we set $\varepsilon_{\text{OT}}=0.05$ and $n_{\text{iter}}=50$.
No additional learned projection is applied before computing similarity; we rely on the backbone's vision tokens and Sinkhorn normalization to provide a stable semantic similarity in $[0,1]$.
The same primitive $\mathrm{sim}(\cdot,\cdot)$ is used for $s_t^{\mathrm{next}}, s_t^{\mathrm{cur}}, s_t^{\mathrm{stay}}$ in main-paper Eq.~(13).

\paragraph{Reward decomposition and default weights.}
Given the three similarities
\begin{equation}
\begin{aligned}
s_t^{\mathrm{next}} &= \mathrm{sim}(\mathbf{F}_{t+1},\hat{\mathbf{F}}^{+}),\\
s_t^{\mathrm{cur}}  &= \mathrm{sim}(\mathbf{F}_{t},\hat{\mathbf{F}}),\\
s_t^{\mathrm{stay}} &= \mathrm{sim}(\mathbf{F}_{t+1},\mathbf{F}_{t}).
\end{aligned}
\end{equation}
we define the auxiliary scalars
\begin{equation}
a_t = s_t^{\mathrm{next}},
p_t = s_t^{\mathrm{next}} - s_t^{\mathrm{cur}},
m_t = 1 - s_t^{\mathrm{stay}},
n_t = [\,\varepsilon - p_t\,]_+,
\end{equation}
with tolerance $\varepsilon > 0$ and $[x]_+=\max(x,0)$.
The shaped reward in main-paper Eq.~(13) is then
\begin{equation}
\label{eq:efn-reward-app}
\begin{aligned}
r_t =\;&
w_{\text{abs}}\, a_t
+ w_{\text{prog}}\, [p_t]_+
+ w_{\text{mot}}\, m_t \\
& -\, w_{\text{lazy}}\,\big(s_t^{\mathrm{next}}\, n_t\, s_t^{\mathrm{stay}}\big)
- \lambda_{\text{time}},
\end{aligned}
\end{equation}
which is finally clipped to $[-1,1]$ for numerical stability.
Unless otherwise stated we use
\begin{equation}
w_{\text{abs}} = 0.1,
w_{\text{prog}} = 20,
w_{\text{mot}} = 0.3,
w_{\text{lazy}} = 20,
\end{equation}
\begin{equation}
\varepsilon = 5\times 10^{-3},\quad
\lambda_{\text{time}} = 0.01.
\end{equation}
These values give a reward distribution with non-trivial variance while keeping $r_t$ in a moderate range for SAC.
In practice, we tune the weights as follows.
We first fix $w_{\text{abs}}$ to a small constant that preserves the absolute similarity bonus without dominating the other terms, then adjust $w_{\text{prog}}$ so that genuine progress toward the retrieved successor frame is reliably rewarded.
If the agent tends to move without approaching the target view, we slightly reduce $w_{\text{prog}}$ and increase $w_{\text{mot}}$.
If the policy learns to idle at viewpoints that already match the target frame, we increase $w_{\text{lazy}}$ (and, if necessary, $\varepsilon$) to strengthen the anti-idling penalty.
The per-step time cost $\lambda_{\text{time}}$ is kept small and is only increased when trajectories become unnecessarily long.
The residual $\ell_2$ penalty from the semantic reward in main-paper Sec.~3.3 is disabled in our final experiments, that is, we set $\lambda_{\text{res}}=0$ and let the tanh-squashed residual range control the magnitude of $\Delta \mathbf{a}_t$.

\paragraph{SAC hyperparameters.}
We use a standard off-policy Soft Actor--Critic setup on top of the shaped reward in Eq.~\eqref{eq:efn-reward-app}.
The discount factor is $\gamma = 0.98$.
The SAC temperature $\alpha$ is adapted online to match a target entropy.
We maintain $\log\alpha$ as a learnable scalar and update it with
\begin{equation}
\mathcal{L}_{\alpha} = -\,\big(\log\alpha\big)\,\big(\log \pi_\phi(\Delta\mathbf{a}_t|\mathbf{c}_t) + \mathcal{H}_{\text{target}}\big),
\end{equation}
where $\mathcal{H}_{\text{target}}$ is proportional to the dimensionality of the residual action.
Since $\Delta\mathbf{a}_t$ lives in a $4\times 4096$ latent space, we set
\begin{equation}
\mathcal{H}_{\text{target}} = -\kappa \cdot (4\cdot 4096), \quad \kappa = 0.25,
\end{equation}
which encourages a moderately stochastic residual policy while still exploiting the shaped reward.

Both actor and critics are trained with Adam optimizers~\cite{kingma2014adam}.
We use learning rates
\begin{equation}
\text{lr}_{\text{actor}} = 3\times 10^{-4},
\text{lr}_{\text{critic}} = 3\times 10^{-4},
\text{lr}_{\alpha} = 3\times 10^{-4},
\end{equation}
with default Adam momentum parameters and no weight decay.
The replay buffer capacity is $2\times 10^5$ transitions, and we start SAC updates once the buffer contains at least one mini-batch.
Each update samples a batch of size $B=32$ and performs one critic update, one actor update, one temperature update, and a soft target update.
The target critic $Q_{\bar{\theta}}$ is updated by Polyak averaging with coefficient
\begin{equation}
\tau = 0.005,\quad
\bar{\theta} \leftarrow (1-\tau)\,\bar{\theta} + \tau\,\theta
\end{equation}
after every gradient step.
The discount $\gamma$ and $\tau$ are kept fixed across all experiments, and we verify that moderate changes around these values have little qualitative effect on EFN's behavior.

\paragraph{Training loop.}
Algorithm~\ref{alg:efn-sac} summarizes the training loop that combines retrieval, reward shaping, and SAC.
We emphasize that the frozen VLA backbone and experience bank are treated as environment-side components; gradients never flow into the backbone or the bank.

\begin{algorithm}[t]
  \caption{EFN training with reward shaping and SAC}
  \label{alg:efn-sac}
  \small
  \begin{algorithmic}[1]
    \State Initialize frozen VLA policy $\pi_0$, experience bank $\mathcal{B}$, residual actor $\pi_\phi$, critics $Q_{\theta_1},Q_{\theta_2}$, target critic $Q_{\bar{\theta}_i}$, temperature $\alpha$, and replay buffer $\mathcal{D}$.
    \For{each training epoch}
      \State Reset environment and obtain initial observation $s_0$; extract vision features $\mathbf{F}_0$ and base action $\mathbf{a}_0^{(0)}$ from $\pi_0$.
      \While{episode not terminated}
        \State Query the experience bank with $(\mathbf{F}_t,\text{instruction})$ to retrieve $(\hat{\mathbf{F}},\hat{\mathbf{a}},\hat{\mathbf{F}}^{+},\ell)$.
        \State Form context $\mathbf{c}_t = \mathrm{enc}(\mathbf{F}_t,\mathbf{a}_t^{(0)},\hat{\mathbf{F}},\hat{\mathbf{a}},\ell)$ and sample residual $\Delta\mathbf{a}_t \sim \pi_\phi(\cdot|\mathbf{c}_t)$.
        \State Execute corrected action $\mathbf{a}_t = \mathbf{a}_t^{(0)} + \Delta\mathbf{a}_t$ in the environment and observe next state $s_{t+1}$, features $\mathbf{F}_{t+1}$, and done flag $d_t$.
        \State Compute similarities
        \[
        s_t^{\mathrm{next}} = \mathrm{sim}(\mathbf{F}_{t+1},\hat{\mathbf{F}}^{+}),\;
        s_t^{\mathrm{cur}} = \mathrm{sim}(\mathbf{F}_{t},\hat{\mathbf{F}}),\;
        s_t^{\mathrm{stay}} = \mathrm{sim}(\mathbf{F}_{t+1},\mathbf{F}_{t}),
        \]
        then construct $a_t,p_t,m_t,n_t$ and reward $r_t$ according to Eq.~\eqref{eq:efn-reward-app}.
        \State Store transition $(\mathbf{c}_t,\Delta\mathbf{a}_t,r_t,\mathbf{c}_{t+1},d_t)$ in $\mathcal{D}$.
        \If{$|\mathcal{D}|\geq B$}
          \State Sample a mini-batch from $\mathcal{D}$ and compute critic targets $y_t$ with the soft Bellman backup using $Q_{\bar{\theta}_i}$ and $\alpha$ as in main-paper Sec.~3.3.
          \State Update critics by minimizing $\mathcal{L}_{\text{critic}}$ and update actor by minimizing $\mathcal{L}_{\text{actor}}$.
          \State Update temperature $\alpha$ toward the target entropy and apply Polyak averaging to refresh $Q_{\bar{\theta}_i}$.
        \EndIf
        \State Set $t\leftarrow t+1$.
      \EndWhile
    \EndFor
  \end{algorithmic}
\end{algorithm}
This procedure realizes the conceptual objective in main-paper Sec.~3.3: EFN learns a residual policy that steers $\mathbf{F}_{t+1}$ toward the retrieved successor frame $\hat{\mathbf{F}}^{+}$ while discouraging degenerate idling and overly long trajectories.

\subsection{EFN Architecture and Implementation}

EFN operates on the latent interface exposed by the frozen vision--language--action (VLA) backbone. At each non-blank step, the backbone provides a set of visual patch embeddings of shape $(256, 4096)$ and a small number of latent action tokens of shape $(4, 4096)$. For the current state and the retrieved experience state, EFN receives both the visual embeddings and the latent tokens and predicts a residual in the same latent space, which is added to the backbone action tokens. All EFN parameters are trained while the backbone remains fixed, so only this residual branch is updated.

To keep the architecture lightweight yet expressive, we first compress the long visual sequence into a few learned latent vectors. This is implemented with a Multihead Attention Pooling (MAP) block inspired by Set Transformers and UniVLA~\cite{bu2025univla}. A bank of learned seed vectors plays the role of latent queries, while the per-patch visual embeddings act as keys and values. A custom attention module projects seeds and visual inputs into multihead query, key and value tensors, performs scaled dot-product attention, and projects the attended seeds back to the embedding dimension. Each MAP block is wrapped with RMS normalization and a position-wise feed-forward network. The feed-forward network uses a SwishGLU unit that applies a linear projection to twice the hidden size, splits it into a value and a gate, modulates the value with a SiLU-activated gate, and maps back to the embedding dimension. This yields compact latent representations of the current and retrieved views instead of operating directly on all $256$ visual patches.

\begin{table}[t]
  \centering
  \small
  \caption{Per-step latency of EFN variants. The full actor--critic configuration is used as reference.}
  \label{tab:efn-latency-ablation}
  \setlength{\tabcolsep}{4pt}
  \begin{adjustbox}{width=\columnwidth}
  \begin{tabular}{lcc}
    \toprule
    Variant & Latency (ms / step)$\downarrow$ & $\Delta$ vs full EFN \\
    \midrule
    policy+critic, 2-layer enc.              & 48.96 & $\color{red}{0.00\%}$ \\
    policy only, 2-layer enc.                & 27.89 & $\color{ForestGreen}{-43.04\%}$ \\
    policy only, 1-layer enc.                & 26.01 & $\color{ForestGreen}{-46.89\%}$  \\
    policy only, 2-layer enc., 1536-d        & 43.48 & $\color{ForestGreen}{-11.21\%}$ \\
    \bottomrule
  \end{tabular}
  \end{adjustbox}
\end{table}

The EFN policy network uses two MAP blocks to obtain visual latents for the current and retrieved states, each with four latent vectors of dimension $1024$. The current and retrieved latent action tokens are projected into the same $1024$-dimensional space. For cross-conditioning, EFN concatenates the pooled current visual latents, pooled retrieved visual latents, and retrieved latent tokens along the sequence dimension, yielding a context sequence of length $12$. The projected current latent tokens serve as queries. A multihead attention layer with layer-normalized queries, keys, and values computes an attended representation of the four query tokens over this $12$-token context. A small feed-forward block with normalization and dropout refines these tokens, and a two-layer Transformer encoder with pre-normalization further processes the four-token sequence to capture interactions among the latent dimensions. A final LayerNorm and a three-layer MLP map each token back to the original $4096$-dimensional latent space, with a $\tanh$ activation and explicit clipping to keep the residual inside a controlled range. The output is a residual tensor of shape $(4, 4096)$ that is added to the backbone latent action tokens. At deployment, we run this policy branch in inference mode and feed the refined latent tokens to the frozen VLA head without any further gradient updates.

For actor--critic training we pair the policy with a critic that shares the same inputs but has a simpler structure. The critic uses MAP blocks with a single latent per view to obtain global visual summaries for the current and retrieved states, and uses mean pooling over the four latent tokens followed by a linear projection to obtain compact token summaries. The four resulting vectors are concatenated into a single feature of size $4 \times 1024$, normalized with a float32 layer normalization for numerical stability under mixed precision, and passed through a multi-layer perceptron that outputs a scalar value. The critic is used only during training to provide value estimates for the Soft Actor--Critic objective; at test time only the policy branch is active. 

To quantify the overhead introduced by EFN itself, we benchmark several variants of the residual module under the same latent interface. Table~\ref{tab:efn-latency-ablation} reports per-step latency for the full actor--critic configuration and for lighter policy-only variants. Removing the critic already reduces latency by more than $40\%$, and a one-layer encoder achieves the lowest cost while preserving the interface, which confirms that EFN adds only a modest runtime overhead on top of the frozen backbone.

\subsection{Simulation Setup and Baseline Configurations}
\label{sec:sim-setup}

We now detail the simulation benchmarks, backbone policies, evaluation protocol, and baseline implementations used in our experiments.

\paragraph{Benchmarks.}
We primarily evaluate on the LIBERO long-horizon manipulation benchmark~\cite{liu2023libero}, which provides language-conditioned tabletop tasks in a simulated kitchen. Following the official protocol, we use the four standard suites: \emph{LIBERO-Spatial}, \emph{LIBERO-Object}, \emph{LIBERO-Goal}, and \emph{LIBERO-Long}. Each suite groups tasks that emphasize spatial reasoning, object-centric manipulation, goal-directed rearrangement, and extended multi-step interactions, respectively. As reported in Table~1 in the main paper, our EFN variants consistently achieve the highest average success rate across all four suites under the same backbone and interaction budget.

\paragraph{Backbone policies.}
Across all experiments, EFN wraps a frozen vision--language--action backbone. For the LIBERO simulations, we use the officially released checkpoints and configurations of \textbf{OpenVLA}~\cite{kim2024openvla} and \textbf{UniVLA}~\cite{bu2025univla} without architectural modifications. For the real-world experiments, we employ the publicly released \textbf{AgiBot GO-1}~\cite{bu2025agibot} controller as our backbone and keep its perception and low-level control stack unchanged. Its performance on LIBERO is reproduced by strictly following the official README instructions (training script, hyperparameters, and evaluation protocol); we do not perform additional tuning specific to our method, and all baselines share the same backbone initialization.

\paragraph{Evaluation protocol.}
Unless otherwise stated, all reported numbers are averaged over \textbf{50 evaluation episodes} per task and \textbf{3 random seeds} (i.e., 150 rollouts per cell in Table~1 in the main paper). During evaluation, policies act deterministically by using the mean action of the underlying stochastic policy (SAC) and the deterministic retrieval rule of each method. An episode is marked as \emph{Success} if the environment-specific success flag is triggered before the task horizon; otherwise it is counted as a failure. The \emph{Step} metric counts the number of control steps from the first \emph{non-blank} action until success or timeout, where we ignore the initial blank warm-up steps (e.g., simulator loading and robot settling) that occur before the policy starts issuing meaningful actions.

Having fixed the benchmarks, backbones, and evaluation protocol, we now describe the baseline families that we compare EFN against. All baselines share the same frozen VLA backbone, experience bank (when applicable), and interaction budget as EFN.

\subsubsection{kNN-RAG (nearest-neighbor retrieval).}
We implement a non-parametric kNN-RAG baseline that controls the agent purely by retrieving and reusing past actions from the experience bank, without learning any residual policy. At each non-blank step, we encode the current observation and language instruction with the same frozen VLA backbone used by all other methods and obtain a joint visual--language key. This key is compared to all step-level keys stored in the experience bank (main-paper Sec.~3.2) using cosine similarity, and we select the single nearest neighbor in this embedding space. The action stored in that retrieved transition is then directly executed as the agent's action at the current step, i.e., the controller simply ``copies what worked before'' for the most similar past state.

In contrast to residual methods, kNN-RAG does not introduce any additional trainable parameters and does not update the backbone or learn an adaptation head; its behavior is fully determined by the fixed VLA encoder, the similarity metric, and the contents of the experience bank. This makes kNN-RAG a strong retrieval-only baseline: it benefits from semantic retrieval over rich past trajectories, but lacks the capacity to adjust or interpolate actions when the retrieved state is only approximately similar to the current situation, exposing the limitations of naive action copying from memory.

\subsubsection{ResAct-style residual RL baseline.}
We further implement a ResAct-style residual RL baseline that adapts the core ideas of ResAct~\cite{liu2025visual}, namely observation-difference features and action residuals, to our VLA setting. To keep the comparison fair, this baseline shares the same frozen vision--language--action backbone and residual RL infrastructure as our other methods. At each time step \(t\), we obtain a visual embedding \(f_{\text{vis}}(o_t)\) from the VLA's visual tokens (e.g., pooled over patches), along with proprioceptive features \(f_{\text{prop}}(x_t)\) and an instruction or goal embedding \(e(\ell)\). In addition, we cache the previous-step visual embedding \(f_{\text{vis}}(o_{t-1})\) and executed action \(a_{t-1}\), and form an observation-difference feature
\begin{equation}
d_t = f_{\text{vis}}(o_t) - f_{\text{vis}}(o_{t-1}),
\end{equation}
optionally passed through a small MLP and LayerNorm for dimensionality reduction and stabilization. The resulting ResAct state is the concatenation
\begin{equation}
s_t^{\text{ResAct}} = [\, f_{\text{vis}}(o_t),\ d_t,\ a_{t-1},\ f_{\text{prop}}(x_t),\ e(\ell)\,],
\end{equation}
which exposes both the absolute scene content and its recent change, together with the previous control signal, to the residual policy.

On top of this state representation, we reuse exactly the same residual RL architecture as in our main baseline. The actor receives \(s_t^{\text{ResAct}}\) and outputs a residual action \(a_t^{\text{res}} \sim \pi_\phi(a \mid s_t^{\text{ResAct}})\), while the critic estimates \(Q_\theta(s_t^{\text{ResAct}}, a_t^{\text{res}})\). The final control command is still defined as an additive refinement to the frozen backbone action,
\begin{equation}
a_t = a_t^{\text{base}} + a_t^{\text{res}},
\end{equation}
so that we modify only the state input to the residual branch rather than changing the action parameterization. Training uses the same Soft Actor--Critic recipe and hyperparameters as our residual baseline (discount factor, learning rates, batch size, target update schedule, replay buffer size, etc.), and relies solely on the environment reward \(r_t^{\text{env}}\) without any retrieval or similarity-based shaping. During both training and evaluation, we maintain the cached pair \((f_{\text{vis}}(o_{t-1}), a_{t-1})\) online (with \(f_{\text{vis}}(o_{-1})\) and \(a_{-1}\) initialized from the first step), ensuring that ResAct operates under the same backbone, data, and interaction budget as our other baselines while explicitly leveraging observation differences and previous actions in its residual policy.

\subsubsection{R2A-style retrieval-augmented baseline}
\label{app:r2a}
We also implement a retrieval-augmented RL baseline inspired by R2A~\cite{goyal2022retrieval} on top of the same frozen VLA backbone as EFN. Because the original R2A is designed for value-based, non-residual control with a specialized dual-process architecture and multiple auxiliary objectives, we do not attempt a bit-level reproduction. Instead, we construct an \emph{R2A-style retrieval-augmented SAC} that preserves the core idea of conditioning the policy and critic on a retrieval-derived context vector, while adapting the design to our continuous-control, residual-on-VLA setting. Throughout this section we explicitly refer to this baseline as an \emph{adapted implementation} and keep all SAC hyper-parameters and backbone interfaces identical to those of EFN for a fair comparison.

State representations and experience storage follow our EFN setup. At each step $t$, we form a state embedding
\begin{equation}
s_t = \big[f_{\text{vis}}(o_t),\, f_{\text{prop}}(x_t),\, e(\ell)\big],
\end{equation}
where $f_{\text{vis}}$ is the pooled visual embedding from the frozen VLA, $f_{\text{prop}}$ encodes proprioceptive signals (joint angles, end-effector pose, etc.), and $e(\ell)$ is the language embedding of the task instruction. We reuse the same step-level experience bank as EFN, but for R2A-style retrieval we additionally precompute keys $k_i = f_{\text{key}}(s_i)$ for stored states $s_i$ using a small MLP. At decision time we compute a query $q_t = f_{\text{query}}(s_t)$ (sharing parameters with $f_{\text{key}}$ in our implementation), perform $k$-nearest-neighbor search over $\{k_i\}$ using cosine similarity, and obtain indices $\{i_1, \dots, i_K\}$. For each neighbor we build a retrieval feature $\tilde{s}_{i_j}$ by concatenating its state, executed action, and scalar reward, and aggregate them with a single-head attention module:
\begin{equation}
\alpha_j = \operatorname{softmax}_j\!\left(\frac{q_t^\top W_a \tilde{s}_{i_j}}{\sqrt{d}}\right),
\qquad
u_t = \sum_{j=1}^K \alpha_j\, W_v \tilde{s}_{i_j},
\end{equation}
which yields the retrieval context $u_t$ that summarizes task-relevant past experience.

On top of the frozen VLA backbone, we train a residual SAC agent that receives both $s_t$ and $u_t$. The backbone produces a base action $a_t^{\text{base}} = \pi_{\text{VLA}}(o_t, \ell)$, while the residual actor outputs $a_t^{\text{res}} \sim \pi_\phi(a \mid s_t, u_t)$, and the executed action is $a_t = a_t^{\text{base}} + a_t^{\text{res}}$. The critic is similarly conditioned on the retrieval context, $Q_\theta(s_t, u_t, a_t)$, and is trained with the standard SAC objective using only the environment reward $r_t^{\text{env}}$; we do not introduce any similarity-based reward terms or additional auxiliary losses. At training time, every sampled transition from the replay buffer recomputes $u_t$ and $u_{t+1}$ via the same retrieval procedure, and at evaluation time we freeze all parameters and keep retrieval active, so that adaptation arises purely from the parametric policy and value functions conditioned on $u_t$. This design captures the central principle of R2A—using retrieved experience as an additional context for decision making—while remaining compatible with our residual continuous-control setting and avoiding the heavy slot-based memory, bidirectional RNNs, and information-bottleneck losses of the original algorithm.

\subsubsection{GC-TTT-style test-time training baseline}
We additionally implement a goal-conditioned test-time training (GC-TTT) baseline adapted to our VLA+residual setting. Concretely, we treat our residual SAC policy (a residual head on top of a frozen VLA backbone) as a goal-conditioned policy and pair it with an offline replay buffer $\mathcal{D}$ collected during pre-training. Each transition in $\mathcal{D}$ stores $(s_t, a_t, r_t, s_{t+1}, g)$ together with the state embedding $z_t = f_{\text{enc}}(s_t)$ and goal embedding $e(g)$ used by our main method. For every episode in $\mathcal{D}$ we also precompute a discounted trajectory return $R_{\text{traj}} = \sum_{t} \gamma^t r_t$ and attach this scalar to all transitions from that episode, which serves as a simple measure of trajectory quality.

At evaluation time, given a new goal $g^\ast$ we perform a single GC-TTT update before running the episode. From the initial observation $o_0$ we obtain the state embedding $z_0 = f_{\text{enc}}(s_0)$ and compute a relevance score for each stored transition $i$ as
\begin{equation}
\mathrm{sim}_i = \cos(z_0, z_i) + \lambda \cos\big(e(g^\ast), e(g_i)\big),
\end{equation}
where $\lambda$ balances state and goal similarity. We first select the top-$M$ transitions by $\mathrm{sim}_i$ to form a relevant subset $\mathcal{D}_{\text{rel}}$, and then rank them by the associated trajectory return $R_{\text{traj}}$. Keeping only the top $q$-th percentile yields a high-quality subset $\mathcal{D}_{\text{good}}(s_0, g^\ast)$ that is both close to the current state--goal pair and drawn from successful trajectories.

We then adapt the residual policy parameters $\phi$ on $\mathcal{D}_{\text{good}}(s_0, g^\ast)$ using a small number $N$ of gradient steps of a behavior-cloning loss,
\begin{equation}
\mathcal{L}_{\text{GC-TTT}}(\phi)
= - \mathbb{E}_{(s_i, a_i) \sim \mathcal{D}_{\text{good}}} \big[ \log \pi_\phi(a_i \mid s_i, g^\ast) \big],
\end{equation}
while keeping the VLA backbone frozen. After these updates, the adapted policy $\pi_{\phi_{\text{adapt}}}$ is used to control the robot for the entire episode. At the end of the episode, we reset $\phi$ back to the pre-trained weights $\phi_0$ and repeat the same procedure for the next evaluation goal. Compared to the original GC-TTT formulation, our implementation is a critic-free variant that relies on trajectory returns instead of a learned value function and only adapts the lightweight residual head. We therefore view this baseline as an ``adapted GC-TTT'' instantiation tailored to our VLA setting rather than an exact reproduction of the original system.

\subsection{Real-World Setup and Task Specifications}
\label{sec:realworld-setup}

\paragraph{Hardware and environment.}
All real-world experiments are conducted on an AgiBot-G1 manipulator mounted next to a fixed tabletop workspace. The arm is controlled in Cartesian space at 10\,Hz using position and gripper commands issued by the frozen VLA backbone (and EFN, when enabled). A calibrated RGB(-D) camera is rigidly mounted above the workspace and looks down at the table and shelf; its images are resized to the same resolution as in simulation and fed directly into the VLA visual encoder without additional preprocessing. The shelf, box, and drawer are anchored at fixed positions in front of the robot, and all objects are restricted to a safe reachable workspace.

\paragraph{Task specifications.}
We instantiate four real-world manipulation tasks that mirror the long-horizon, language-conditioned scenarios used in simulation: \textbf{BottlePlace}, \textbf{ShelfSort}, \textbf{StockLift}, and \textbf{DrawerStore}. Each task is defined by (i) an initial object configuration, (ii) a target region, and (iii) a success condition.

\noindent \textbf{BottlePlace.}
A plastic bottle is placed on the tabletop within a fixed start region in front of the right arm. A small open box is placed to the side of the table and is treated as the goal region. The task is considered successful if the bottle ends in the interior of the box (its 3D center lies inside the pre-defined box region) and the right gripper is fully opened at the end of the episode. A typical language command is:
\emph{``Use your right arm to pick up the bottle and place it into the box.''}

\noindent \textbf{ShelfSort.}
Several drink containers (e.g., cans and bottles with different labels) are arranged on a tabletop shelf, with at least one item of the same category already present in a designated ``cluster'' region. Another drink from the same category is placed on the shelf in a randomized start position. The success condition requires that the target drink is placed within the cluster region next to visually similar items (within a fixed 3D tolerance) and that the right gripper is released. An example language instruction is:
\emph{``Grasp the drink from the shelf with your right arm and place it next to the similar drinks, then open the gripper.''}

\noindent \textbf{StockLift.}
A stack of identical cans is placed on a small auxiliary tabletop, and a separate shelf contains an empty spot marked as the goal region. The top can of the stack is always designated as the object to manipulate. The task is successful if the robot lifts the top can from the stack, moves it to the shelf, and leaves it resting within the predefined goal region with the right gripper open. A representative command is:
\emph{``Pick up the top can from the small table with your right arm and place it onto the shelf, then release.''}

\noindent \textbf{DrawerStore.}
A packet of napkins is placed on the main tabletop within reach of the right arm. A drawer in front of the robot starts in the closed position and serves as the storage location. The robot must (1) use the left arm to pull the drawer open beyond a minimum displacement; (2) use the right arm to grasp the napkin packet from the tabletop; (3) transfer the packet from the right gripper to the left gripper; and (4) place the packet into the drawer interior. The episode is marked as successful if, at termination, the napkins lie fully inside the drawer volume and both grippers are open. An example language instruction is:
\emph{``Open the drawer with your left arm, pick up the napkin packet with your right arm, hand it to the left arm, and place it into the drawer.''}

\paragraph{Per-episode protocol and experience logging.}
For all four tasks, each episode begins from a manually reset initial configuration with objects placed in their designated start regions and the arms in a neutral pose. At each control step, the overhead camera image and the language instruction are passed through the frozen VLA backbone to produce an action proposal; EFN (or other baselines) then optionally modifies this proposal before it is sent to the low-level controller at 10\,Hz. We cap the horizon at a fixed maximum number of steps (typically 60--80, depending on the task); if the success condition is met earlier, the episode terminates immediately. On failure (timeout, dropped object, or clear deviation from the goal), the episode is terminated and the environment is manually reset.

During deployment with EFN, we maintain an online experience bank that stores real-world rollouts. For each successful episode, we record the RGB images, language instruction, latent visual embeddings, latent action tokens, and executed actions at every non-blank step, and append these transitions to the experience bank. Failed or aborted episodes are not written to the bank by default to avoid contaminating the retrieval database with suboptimal behaviors.

\paragraph{Safety considerations.}
All experiments are run with conservative joint and workspace limits, velocity and torque bounds, and soft joint limits enabled on the AgiBot-G1 controller. Episodes are immediately stopped and reset if any potential collision or unsafe configuration is detected. These safety measures slightly reduce the usable workspace but do not otherwise affect the task definitions or evaluation protocol described above.

\section{Rationale behind Design}
\label{rationale}
\subsection{Design Objectives and Constraints}
\label{sec:design-objectives}

Our starting point is a practical limitation of current vision--language--action (VLA) policies: once trained and deployed, the backbone is typically frozen and cannot \emph{learn from new experience} collected in the target environment. However, real-world deployment quickly exposes distribution shift, long-horizon credit assignment issues, and subtle failure modes that are hard to anticipate in advance. 

Our primary objective is therefore to endow a \emph{frozen} VLA backbone with a mechanism that can still ``become smarter'' over time at deployment, improving success rate and reducing action steps by exploiting past executions, without ever modifying backbone weights.
This objective is shaped by several hard constraints arising from realistic deployment scenarios:
\begin{itemize}[leftmargin=*,topsep=0pt]
    \item \textbf{No backbone finetuning.} The core VLA policy is treated as a fixed, pre-trained component. We do not update its parameters during deployment; all adaptation must occur through an auxiliary module that interfaces with its latent representations.
    \item \textbf{Limited compute at deployment.} Real robots and embedded platforms cannot afford frequent, large-scale finetuning or replay-based training. Any learning or adaptation mechanism must be lightweight, incremental, and compatible with strict latency and memory budgets.
    \item \textbf{Safety and reproducibility.} Deployment-time changes must be controlled and auditable. Residual corrections should stay in a safe regime (e.g., not arbitrarily overriding the backbone), and the overall procedure should remain stable enough to reproduce results across runs and hardware setups.
    \item \textbf{Backbone-agnostic interface.} The adaptation mechanism should generalize across different VLA architectures (e.g., OpenVLA, UniVLA, AgiBot-G1 controllers) without requiring backbone-specific surgery. This motivates operating purely on the shared latent interface (visual tokens, language tokens, and latent action tokens) exposed by the frozen policy.
    \item \textbf{Minimal code and system changes.} To ease integration into existing stacks, we aim to attach a thin ``add-on'' module on top of the VLA's latent action interface, rather than redesigning perception, language processing, or low-level control.
\end{itemize}

All subsequent design choices in our framework, including experience storage, retrieval, residual policy architecture, and training objectives, are made to satisfy these constraints while serving the central goal of \emph{deployment-time improvement with a frozen backbone}. The proposed Experience Feedback Network (EFN) should therefore be viewed as a minimal, safe, and portable layer that augments a unified VLA policy with post-deployment learning capabilities, rather than a replacement for the underlying model.

\subsection{Why a Frozen VLA with a Residual Head?}
\label{sec:why-residual}

Given the objectives and constraints in Sec.~\ref{sec:design-objectives}, we deliberately start from a \emph{frozen} vision--language--action (VLA) backbone and attach a light residual head, instead of (i) directly finetuning the large model at deployment, or (ii) retraining a new policy from pixels to actions.

Continuing to finetune the backbone online is conceptually simple but practically fragile. It requires non-trivial compute and memory on the robot, frequent gradient updates, and careful replay management to avoid catastrophic forgetting. More importantly, aggressive finetuning risks damaging the strong cross-task generalization that large VLAs have already acquired on benchmarks such as LIBERO and our real-world AgiBot-G1 tasks. Once the backbone drifts, it is difficult to guarantee reproducibility across deployments and hardware.

On the other hand, discarding the VLA and retraining a full pixel-to-action policy would forfeit the benefits of large-scale pre-training altogether. Such a design would be data-inefficient, slow to adapt in deployment, and tightly coupled to a specific robot and environment, contradicting our goal of a backbone-agnostic, reusable mechanism.

A residual head offers a middle ground. We keep the base policy’s action $a_t^{\text{base}}$ as the default behavior and learn only a small correction $\Delta a_t$:
\begin{equation}
a_t = a_t^{\text{base}} + \Delta a_t.
\end{equation}
This preserves the backbone’s strong generalization and task coverage, while allowing EFN to refine local decisions where the base policy systematically underperforms. Because the residual only needs to model \emph{deviations} from a competent policy, it is more sample-efficient and typically more stable to train, consistent with the intuition behind residual policy learning and ResAct-style methods~\cite{liu2025visual}.

Our ablations further support this design. When we replace the residual head with a standalone action predictor that ignores the base action (``no residual / direct action prediction''), performance drops in both success rate and step efficiency (see Appendix Section~\ref{app:direct-action}). In other words, throwing away the base action and asking the small head to solve the entire control problem again is strictly worse than learning corrections on top of it.

Compared to prior residual RL baselines such as ResAct, EFN’s residual is \emph{retrieval-driven} rather than blind: the correction $\Delta a_t$ is conditioned on the current latent state and on retrieved experience that encodes how similar situations were handled in the past. This retrieval-guided residual design is central to EFN: it lets us keep the backbone frozen, leverage its existing generalization, and still obtain deployment-time improvements through small, data-efficient adjustments.

\subsection{Why an Episodic Experience Bank?}
\label{sec:why-episodic-bank}

To support deployment-time improvement with a frozen backbone, we organize past executions into an \emph{episodic} experience bank. Instead of storing isolated transitions, we keep complete rollouts $\tau = \{(o_t, a_t, \dots)\}_{t=1}^T$ together with their language instruction and outcome. At the \emph{trajectory} level, we encode the instruction into a fixed-dimensional embedding $\ell_\tau$, which is used for instruction-filtered retrieval during deployment. At the \emph{step} level, we store compact visual features, retrieval keys, and the corresponding base actions. This two-level structure lets us first select a small set of candidate episodes that match the current instruction, and then perform fine-grained, step-wise retrieval of locally similar states within those episodes.

The retrieval key for each step combines mean and max pooled features. Concretely, given a set of visual tokens $\{v_i\}_{i=1}^N$, we compute
\begin{align}
\mu &= \frac{1}{N} \sum_{i} v_i, \\
m &= \max_{i} v_i \quad (\text{element-wise}), \\
k &= \operatorname{LN}([\mu; m]),
\end{align}
and L2-normalize $k$ before computing cosine similarity. Intuitively, the mean component captures the global scene configuration, while the max component emphasizes salient objects and local cues that may be critical for manipulation. L2-normalization followed by cosine similarity yields a scale-insensitive, numerically stable similarity measure across different backbones and environments. In practice, we found this mean–max fusion more robust than simple global average pooling alone, which tended to produce noisier neighbors and more sensitivity to background clutter.

We further impose two simple but important priors in retrieval: \emph{instruction filtering} and an \emph{efficiency prior}. Instruction filtering uses $\ell_\tau$ to restrict retrieval to episodes whose language embedding is close to the current instruction, which reduces harmful cross-task transfer on multi-task benchmarks such as LIBERO and prevents the agent from imitating “good actions for the wrong task.” The efficiency prior biases the bank towards shorter successful trajectories: during offline construction and online retrieval, we prefer episodes that reach success in fewer steps. This aligns with deployment-time goals (reusing concise, high-quality behaviors) and empirically improves both success and step efficiency; removing instruction filtering or the length prior (“w/o instruction embed”) degrades performance in our ablations (see Appendix Table~\ref{tab:ab}).

Finally, we emphasize that the experience bank and its retrieval interface are shared across all retrieval-based baselines. kNN-RAG, R2A-style, GC-TTT-style methods, and EFN all operate on the \emph{same} episodic bank, use the \emph{same} key/query construction and cosine similarity, and differ only in how they \emph{consume} the retrieved neighbors (direct action copying, test-time gradient updates, or retrieval-conditioned residuals). This shared interface ensures that EFN does not benefit from any hidden advantage at the retrieval level; improvements arise solely from how retrieved experience is integrated into action selection.

\begin{figure*}[t]
    \centering
    \begin{subfigure}{0.49\linewidth}
        \centering
        \includegraphics[width=\linewidth]{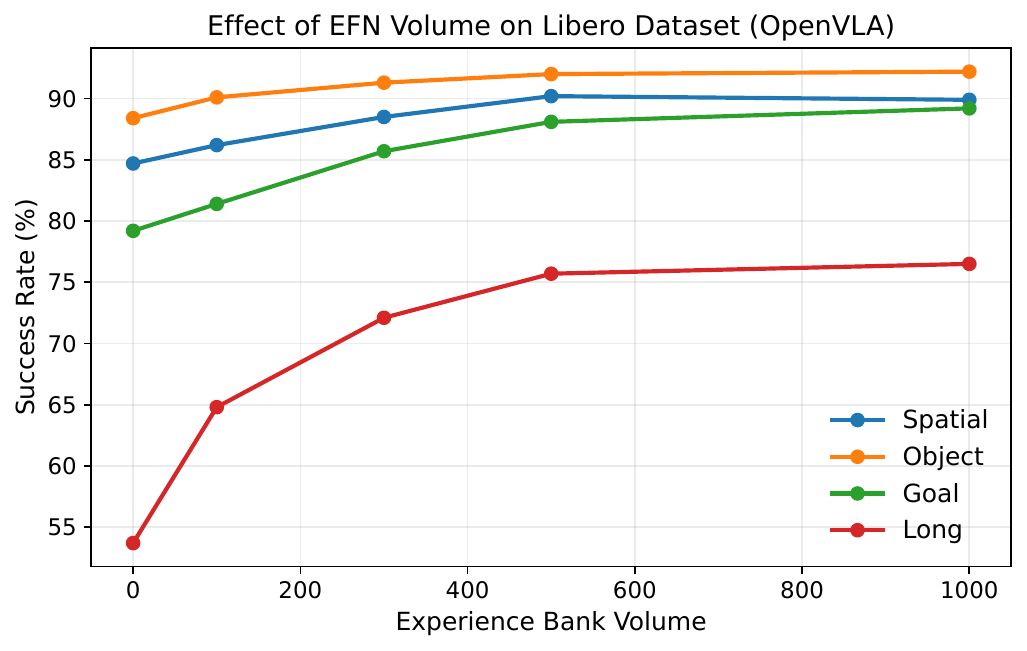}
        \caption{LIBERO tasks}
        \label{fig:libero}
    \end{subfigure}
    \hfill
    \begin{subfigure}{0.49\linewidth}
        \centering
        \includegraphics[width=\linewidth]{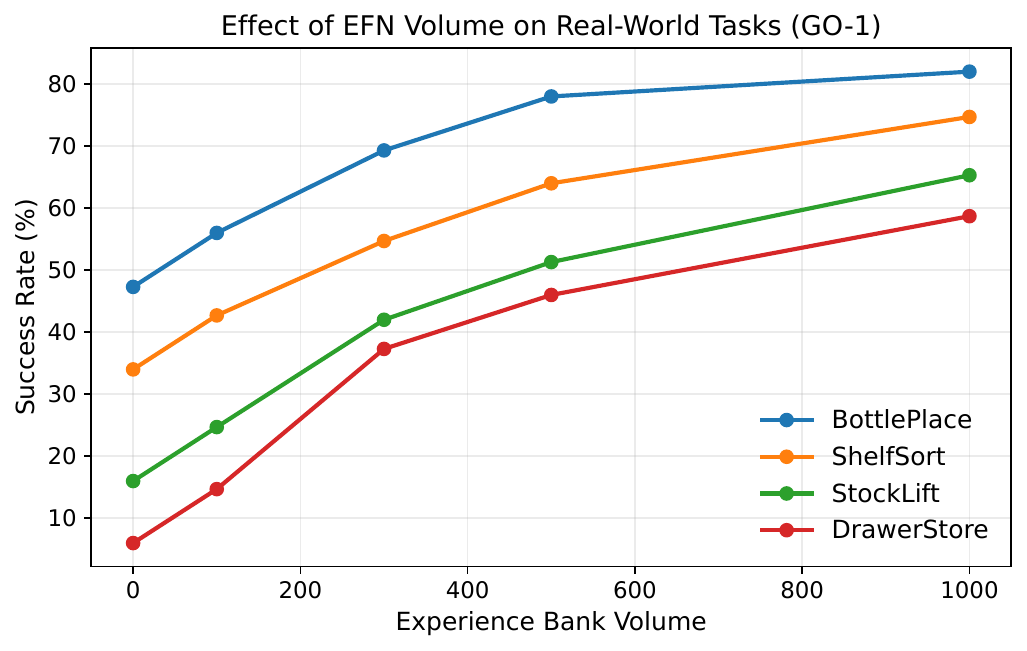}
        \caption{Real-world tasks}
        \label{fig:go1}
    \end{subfigure}
    
    \caption{Performance consistently improves as the experience bank grows in both simulated LIBERO tasks with OpenVLA and real-world manipulation tasks with GO-1. Here, the bank volume refers to the number of stored trajectories (episodes) in the experience bank.}
    \label{fig:threeplots}
\end{figure*}

\subsection{Why Residual Policy Optimization Instead of Direct Action Prediction?}
\label{sec:why-residual-optimization}

From an optimization perspective, directly predicting actions with a new lightweight head is equivalent to discarding the base policy altogether. The new actor must re-learn the full mapping from latent state to action, effectively reconstructing the entire action manifold of the task. For a small network trained from sparse, delayed rewards at deployment, this is a hard problem: the search space is large, gradients are noisy, and the sample complexity is high. In practice, such a head tends to overfit to a small set of trajectories and fails to reliably generalize across the diverse scenes and tasks that the frozen VLA backbone already handles well.

Residual policy optimization takes a different view. Instead of learning a fresh action $a_t^{\text{new}}$, EFN learns a correction $\Delta a_t$ around the base action $a_t^{\text{base}}$:
\begin{equation}
a_t \;=\; a_t^{\text{base}} + \Delta a_t.
\end{equation}
This constrains the RL search space to a local neighborhood of $a_t^{\text{base}}$, which is already a strong, pre-trained solution. The residual head only needs to model how to adjust actions in cases where the backbone is systematically biased or underspecified, rather than solving the whole control problem from scratch. This local parameterization smooths the optimization landscape, reduces the effective dimensionality of the problem, and makes gradient-based updates more stable and data-efficient. Moreover, when retrieval is unreliable or EFN is uncertain, the learned residual naturally shrinks towards zero, and the agent safely falls back to the base action $a_t^{\text{base}}$.

Our ablations corroborate this reasoning: a variant that replaces the residual optimization with direct action prediction (ignoring $a_t^{\text{base}}$) consistently underperforms EFN in both success and step metrics (see Sec.~\ref{exp} and the ``direct action prediction'' rows in the supplementary tables). These results suggest that, under a frozen VLA backbone, optimizing a retrieval-conditioned \emph{residual} is a strictly more favorable formulation than learning a standalone action head.

\subsection{Why Similarity-shaped Dense Rewards and the Anti-idling Design?}
\label{sec:why-sim-reward}

The reward decomposition in main-paper Fig.~4 is designed to turn retrieval into a dense and behaviorally aligned learning signal, rather than relying solely on sparse task success. A naive starting point is to reward the agent by how well its next observation matches the retrieved next frame, e.g., a cosine similarity term $a_t \in [0,1]$ between the actual next visual embedding and the retrieved target. This already provides a dense signal at every step, but it also introduces a failure mode: the agent can exploit the reward by finding a visually similar pose and then ``vibrating in place,'' maintaining high similarity without making real progress toward task completion.

To avoid this, we explicitly split the similarity-shaped reward into components that distinguish \emph{progress} from \emph{stagnation}. The progress term $[p_t]_+$ measures how much closer the agent moves towards the retrieved next frame compared to the previous step, and only positive progress is rewarded. This encourages trajectories that steadily approach the retrieved outcome, rather than merely staying close once a high-similarity configuration is found. In addition, we introduce a motion term $m_t$ and an anti-idling penalty weighted by $w_{\text{lazy}}$: steps that change the scene or end-effector state in a meaningful way are rewarded, whereas near-static steps (high similarity but negligible motion) are penalized. Finally, a temporal coefficient $\lambda_{\text{time}}$ discourages unnecessarily long trajectories, biasing the policy towards concise, decisive behaviors that reach success in fewer steps.

From this perspective, the reward is not a black-box signal delegated entirely to SAC to ``figure out'' credit assignment; instead, it directly encodes our behavioral preferences: (i) match good retrieved futures, (ii) \emph{move toward} them rather than hovering idly nearby, and (iii) reach them quickly. The ablations in Table~\ref{tab:ab} (also summarized in main-paper Fig.~4) support this design. Removing the similarity-shaped dense terms (``w/o dense rewards'') and relying only on sparse task success leads to lower average success and less consistent improvements over OpenVLA, indicating that the dense retrieval-based components are crucial for effective post-deployment learning. Dropping the anti-idle terms (``w/o anti-idle'') yields a policy that attains similar success but with noticeably longer episodes, confirming that, without explicit penalties on idleness, the agent tends to favor visually safe but inefficient behaviors. In contrast, the full EFN reward decomposes the similarity signal into progress, motion, anti-idle, and time components, producing the best overall trade-off between success and step efficiency on LIBERO.

\subsection{Why Reinforcement Learning for EFN?}
\label{sec:why-rl}

A natural question is why EFN is trained with reinforcement learning rather than supervised learning. The core reason is that our training signal is defined \emph{through} the environment dynamics and therefore does not admit straightforward backpropagation. As described in Sec.~3.3, after executing $\mathbf{a}_t$ the environment advances to a new state $s_{t+1}$ with vision features $\mathbf{F}_{t+1}$. We then define a dense semantic similarity reward
\begin{equation}
r^{\text{sem}}_t \;=\; \cos\!\big(\mathbf{u}(\mathbf{F}_{t+1}),\, \mathbf{u}(\hat{\mathbf{F}}^{+})\big),
\end{equation}
where $\hat{\mathbf{F}}^{+}$ is the retrieved next-frame feature and $\mathbf{u}(\cdot)$ is a projection into the similarity space. Both $s_{t+1}$ and $\mathbf{F}_{t+1}$ are only revealed after interacting with the simulator or real robot. Since the transition $s_t \xrightarrow{\mathbf{a}_t} s_{t+1}$ is non-differentiable and we do not have a tractable model of the environment, we cannot propagate gradients from $r^{\text{sem}}_t$ back through the dynamics to EFN parameters as in standard supervised learning. Instead, we are precisely in the canonical RL setting: EFN chooses actions, observes next states and rewards from a black-box environment, and must improve its policy from these interaction signals.

Within RL, we adopt Soft Actor--Critic (SAC) for training EFN. First, our control space is continuous (especially on the real AgiBot-G1 platform), making SAC a natural choice as a widely used off-policy algorithm for continuous actions. Second, the off-policy nature of SAC lets us aggressively reuse transitions from the experience bank and the ongoing rollouts, which aligns with our central theme of ``learning from experience'' rather than discarding past data. Third, SAC’s entropy regularization is particularly helpful in our residual setting: it encourages exploration instead of collapsing too early to the trivial solution $\Delta \mathbf{a}_t \approx \mathbf{0}$, and stabilizes learning when the dense similarity-shaped rewards and anti-idle penalties are combined.

Our ablations support these design decisions. When we remove SAC and rely only on a value-style critic without the entropy-regularized actor update (“w/o SAC’’ in Table~\ref{tab:ab}), EFN’s improvements over the backbone shrink in both success and step metrics. In contrast, the full SAC-based optimization leverages the dense similarity rewards and the retrieval structure to deliver consistent gains on top of a frozen VLA policy. Overall, reinforcement learning—and SAC in particular—fits our design philosophy: a lightweight optimization layer that operates on the VLA’s latent interface, reuses interaction experience, and learns directly from deployment-time feedback without modifying any backbone weights.

\begin{figure*}[ht]
    \centering
    \begin{minipage}{0.23\textwidth}
        \centering
        \includegraphics[height=3.05cm]{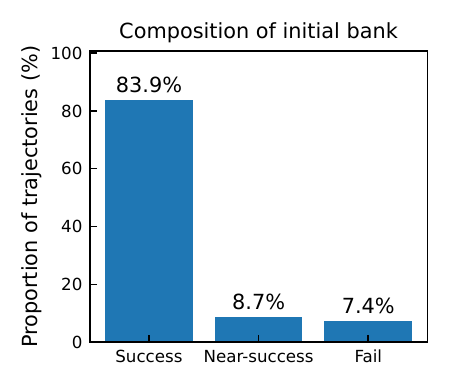}  
        \caption{Composition of successful, near-successful, and failed trajectories on LIBERO.}
        \label{fig:experience_bank_composition}
    \end{minipage}\hfill
    \begin{minipage}{0.23\textwidth}
        \centering
        \includegraphics[height=3.05cm]{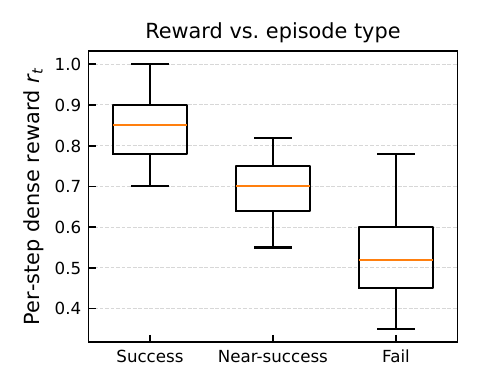}  
        \caption{Per-step reward distribution for successful, near-successful, and failed episodes.}
        \label{fig:reward_distribution}
    \end{minipage}\hfill
    \begin{minipage}{0.23\textwidth}
        \centering
        \includegraphics[height=3.05cm]{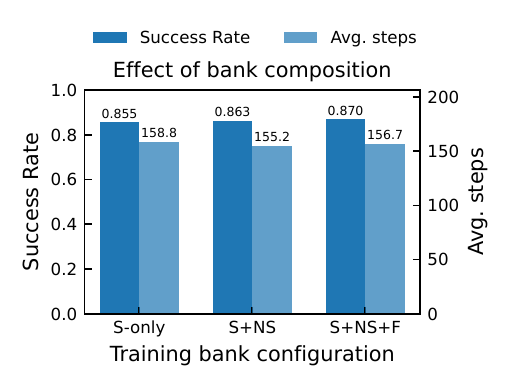}  
        \caption{Training performance with different experience bank compositions on LIBERO.}
        \label{fig:ablation_study}
    \end{minipage}\hfill
    \begin{minipage}{0.23\textwidth}
        \centering
        \includegraphics[height=3.05cm]{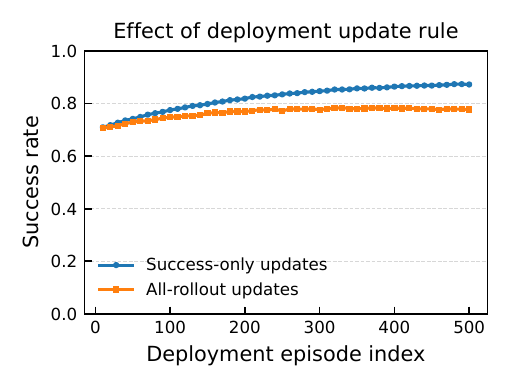}  
        \caption{Comparison of rollout update strategies during deployment: Success-only vs All.}
        \label{fig:deployment_update_strategy}
    \end{minipage}
\end{figure*}

\subsection{Why Live Experience Growth at Deployment?}

Our deployment-time design explicitly separates \emph{what can change} (the experience bank) from \emph{what remains fixed} (the VLA backbone and EFN weights), so that the system can adapt in the field without sacrificing stability.

\paragraph{Why not update network parameters at deployment?}
Updating parameters online turns deployment into test-time training (TTT), which easily leads to instability and hard-to-debug performance collapse; our GC-TTT baseline empirically illustrates this risk. Moreover, backpropagating through a large VLA in the wild is compute- and engineering-heavy, and makes it difficult to guarantee latency and reliability. Keeping the backbone and EFN frozen also simplifies reproducibility: a fixed set of weights yields consistent behavior, whereas a continuously updated model is a moving target whose behavior depends on its entire online training history.

\paragraph{Why still grow the experience bank?}
Freezing weights does not preclude “learning from experience”: EFN adapts by changing \emph{which} trajectories it can retrieve, not \emph{how} it processes them. As shown in Figure~\ref{fig:threeplots}, enlarging the experience bank consistently improves success rates on LIBERO and real-world GO-1 tasks, because a larger bank provides more relevant and diverse execution snippets for residual guidance on top of the frozen VLA. By appending only validated (successful) rollouts, this live growth remains a safe form of adaptation: the agent gains from accumulating high-quality experiences while preserving the stability, compute efficiency, and auditability of a frozen model.

\section{Discussion, Limitations, and Future}
\label{dis}

\subsection{Success and Failure Experience}
\label{sec:success-failure}

A key design choice in EFN is to treat success, near-success, and failure differently at \emph{training} and \emph{deployment} time. During training, the offline bank intentionally contains a mixture of successful rollouts, near-successful rollouts that complete only a subset of sub-goals, and even clearly failed trajectories (Figure~\ref{fig:experience_bank_composition}). This is feasible because our reward $r_t$ is defined purely from semantic feature similarity rather than a binary success flag: any transition that moves the state closer to a goal-consistent embedding can provide useful signal, regardless of whether the episode ultimately succeeds.

Figures~\ref{fig:reward_distribution} and~\ref{fig:ablation_study} support this design. The per-step reward distributions show that failed and near-successful episodes still contain many high-similarity steps, i.e., meaningful partial progress. In the ablation on bank composition, moving from \texttt{S-only} to \texttt{S+NS} consistently improves performance, and even \texttt{S+NS+F} does not significantly degrade results, indicating that EFN is robust to a moderate amount of failure noise when the reward is similarity-based.

In contrast, during deployment we only append rollouts classified as successful to the live bank. Success detection is fully automatic: instead of human labels, we reuse the same feature-matching criterion as Eq.~(11) in the main paper and mark an episode as successful if its final state embedding is sufficiently similar to the goal embedding. Figure~\ref{fig:deployment_update_strategy} shows that appending all rollouts (\texttt{All-rollouts}) gradually harms performance, whereas a success-only update strategy maintains a cleaner bank and yields higher, more stable success rates. A remaining limitation is that our current detector is a simple similarity threshold; future work could explore learned, task-aware success estimators or soft weighting of borderline near-successes to further refine which experiences are admitted into the live bank.

\begin{table*}[h]
\centering
\caption{Ablation studies on the LIBERO benchmark}
\begin{adjustbox}{max width=\linewidth}
\begin{tabular}{cllllllllll}
\hline
 & \multicolumn{2}{c}{Spatial} & \multicolumn{2}{c}{Object} & \multicolumn{2}{c}{Goal} & \multicolumn{2}{c}{Long} & \multicolumn{2}{c}{Average} \\ 
\multirow{-2}{*}{Method} & Succ. & Step & Succ. & Step & Succ. & Step & Succ. & Step & Succ. & Step \\ \hline
OpenVLA & 84.7 & 119.5 & 88.4 & 163.7 & 79.2 & 121.5 & 53.7 & 275.9 & 76.5 & 160.2 \\
w/o SAC & 80.4 & 124.2 & 81.5 & 170.9 & 76.7 & 122.3 & 42.6 & 282.3 & 70.3 & 161.2 \\
w/o dense rewards & 85.6 & 117.0 & 90.1 & 161.4 & 84.3 & 119.4 & 66.8 & 270.9 & 81.7 & 161.3 \\
w/o instruction embed & 88.1 & 114.2 & 90.7 & 162.1 & 86.9 & 116.8 & 74.4 & \textbf{262.1} & 85.0 & 160.0 \\
w/o anti-idle & 89.9 & 120.9 & 91.3 & 162.9 & \textbf{88.2} & 123.1 & 75.4 & 280.2 & 86.2 & 167.4 \\
OpenVLA+EFN(full) & \textbf{90.2} & \textbf{111.8} & \textbf{92.0} & \textbf{158.3} & 88.1 & \textbf{114.5} & \textbf{75.7} & 264.3 & \textbf{86.5} & \textbf{158.2} \\ \bottomrule
\end{tabular}
\end{adjustbox}
\label{tab:ab}
\end{table*}

\subsection{Experience-centric post-deployment learning}

Most prior approaches to post-deployment adaptation in embodied agents are implicitly \emph{parameter-centric}: they treat the agent's weights as the primary locus of change, and focus on finetuning, test-time training, or meta-RL updates to continually reshape the backbone policy. In contrast, EFN adopts an explicitly \emph{experience-centric} view. Once deployed, the weights of both the VLA backbone and the EFN residual controller are frozen; the only component that continues to evolve is the experience bank. Under this perspective, learning after deployment is expressed not as further gradient descent in parameter space, but as the incremental accumulation, organization, and retrieval of interaction histories that remain external to the backbone.

Our empirical comparisons support this reframing. Methods such as R2A-style retrieval-augmented RL and GC-TTT represent the ``keep updating the backbone'' route: they can improve performance, but at the cost of substantial online optimization, sensitivity to hyperparameters, and potential instability when applied to long-horizon manipulation on LIBERO or to real-world AgiBot-G1 deployment. By contrast, EFN uses a lightweight residual head together with retrieval over a growing bank and already achieves comparable or better gains, while leaving the underlying VLA unchanged. This suggests that much of the benefit commonly attributed to continual finetuning can instead be harvested by enriching and exploiting a structured memory of past executions, without repeatedly pushing the base policy into unexplored regions of parameter space.

Viewing post-deployment adaptation as experience growth rather than perpetual finetuning has broader implications. It makes behavior easier to reproduce and audit, since changes over time are driven by the content of an explicit memory store rather than opaque weight drift, and it aligns better with realistic deployment constraints where on-board compute and engineering budgets limit extensive online training. At the same time, it shifts the core challenges toward curating, scaling, and querying experience banks, for example by addressing long-term storage, retrieval efficiency, and coverage of rare but safety-critical events. A more detailed rationale for each architectural choice underlying this experience-centric design is given in Appendix Section~\ref{rationale}; here we emphasize the conceptual shift from parameter-centric to experience-centric post-deployment learning.

\subsection{When does EFN help, and why?}

Empirically, EFN helps most once the experience bank covers a modest set of \emph{typical} situations rather than an extremely large corpus. Increasing the bank volume from $0$ to $300$ episodes already yields a major jump in performance, while further expanding to $1000$ provides only diminishing but consistent gains (Figure~\ref{fig:threeplots}). This suggests that EFN is mainly sensitive to whether common layouts, viewpoints, and language instructions are represented, rather than to exhaustively storing every trajectory. In benchmarks like LIBERO, where tasks share similar manipulation structure, even a medium-sized bank is enough for retrieval to regularly surface useful precedents.

Ablations also clarify what kind of signal is needed. A retrieval-only kNN-RAG controller that directly replays the stored action of the nearest state often hurts performance, indicating that “copying the past” is too brittle under small pose or occlusion shifts. A residual-only variant (ResAct) that ignores episodic retrieval but learns a global residual head yields modest but stable gains, consistent with correcting systematic biases of the base VLA. EFN outperforms both by combining episodic guidance with local adaptation: retrieved trajectories provide a high-level prior, while the residual adjusts to the current observation. On the real AgiBot-G1 tasks, the gap between $300$ and $1000$ episodes is even smaller, hinting that in real deployment the task distribution is more repetitive and sensor noise higher, so memory \emph{quality} matters more than sheer size. We also observe clear failure modes: if the scene is structurally reconfigured (e.g., the tabletop is completely rearranged), old experiences can become misleading, and when the base VLA is very weak, EFN can only provide small local corrections rather than “reviving” the policy from scratch.

\subsection{Limitations}

The first limitation of EFN is its reliance on the backbone representation quality. EFN operates entirely in the latent space exposed by the frozen VLA; if the backbone fails catastrophically for certain objects, viewpoints, or language instructions, both retrieval and residual correction will fail jointly. Moreover, the current design needs a non-trivial number of successful trajectories before it can boost performance: in extremely cold-start regimes with virtually no successes, EFN has little useful signal to retrieve. In practice, we observe that once there are even a handful of successful rollouts, EFN can already yield substantial gains. For example, on the challenging \emph{DrawerStore} task in the real-world AgiBot-G1 experiments, whose base success rate is only \textbf{5.3\%}, EFN increases performance to \textbf{37.3\%} with a bank volume of 300 and further to \textbf{58.7\%} with volume 1000 (Table~2 in the main paper; visualizations in Section~\ref{vis}). Nevertheless, if the backbone truly never succeeds (e.g., close to $0\%$ success), our method provides limited leverage, and the more appropriate remedy is to retrain or replace the underlying VLA rather than rely on post-deployment experience alone.

The second limitation concerns the size and management of the experience bank. Under the scales considered in this work (LIBERO simulation and our AgiBot-G1 experiments), the memory footprint and retrieval latency remain modest, and EFN adds only a small overhead on top of the base VLA policy, as summarized in Table~3 in the main paper. However, in long-running deployments with many robots or continuously operating fleets, the bank will grow without bound unless additional mechanisms are introduced. In such settings, it will be necessary to design more sophisticated caching, compression, and replacement strategies (e.g., task-aware coreset selection, time- or novelty-based forgetting, or hierarchical multi-bank structures) to keep both storage and retrieval cost under control while preserving the benefits of rich episodic experience.

\subsection{Future directions}

A natural next step is to strengthen the \emph{experience interface} itself. In our current design, EFN compares the latent encodings of the present observation against stored experiences, and conditions residual actions on the retrieved matches. An intriguing extension is to couple EFN with world models or predictive representations, so that retrieval can be performed in the space of \emph{future} trajectories rather than only current views. Concretely, one could predict a short-horizon rollout under the base policy, and retrieve experiences whose \emph{successor segments} best align with this predicted future, allowing EFN to correct course before errors manifest. Such predictive retrieval opens the door to uncertainty-aware guidance and more robust long-horizon adaptation.

Another promising direction is to move beyond a per-backbone, per-domain experience bank toward \emph{shared} experience repositories across tasks and robots. In the present work, each VLA backbone maintains its own experience bank, tailored to a single domain or embodiment; this simplifies training but prevents reusing experience across, for example, OpenVLA and UniVLA or between different manipulation platforms. Future work could investigate unified latent representation spaces or lightweight alignment layers that enable cross-robot retrieval and reuse of experience, while guarding against negative transfer. This would turn EFN-style mechanisms into a more general infrastructure for pooling and distilling experience across heterogeneous agents and environments, rather than a separate add-on for each backbone.

\section{Supplementary Experimental Results}
\label{exp}
\subsection{Runtime and Memory Footprint of EFN}
\label{app:efn-runtime-memory}

\subsubsection{Standalone Cost}
We first measure the computational footprint of EFN in isolation, excluding the VLA backbone and the critic. We instantiate the EFN policy using the same configuration as in our main experiments (1024-dimensional embedding, 2-layer Transformer encoder), and feed synthetic latent and visual features that match deployment-time shapes: $(1,4,4096)$ latent tokens and $(1,256,4096)$ visual embeddings.

\begin{table}[h]
\centering
\caption{Standalone runtime and memory footprint.}
\label{tab:efn-overhead}
\begin{tabular}{lc}
\toprule
Method & Bare EFN policy \\
\midrule
Runtime (ms / step)      & 1.944 \\
Param memory (MB)        & 208.17 \\
Peak activations (MB)    & 2.59 \\
\bottomrule
\end{tabular}
\end{table}

We benchmark $200$ forward passes after $50$ warm-up iterations using CUDA event timing. As shown in Table~\ref{tab:efn-overhead}, EFN adds only $\sim\!1.94$\,ms per control step and incurs minimal activation memory ($\sim\!2.6$\,MB). The parameter footprint is modest ($208$\,MB), especially compared to the frozen VLA backbone. These results indicate that EFN is lightweight enough for real-time control and can be cleanly integrated into existing policies.

\subsubsection{Integration Cost}
\label{cost}
We next examine the end-to-end overhead when EFN is integrated into the full GO-1 policy, including vision encoding, LLM inference, and action decoding. This measurement in Table~\ref{tab:efn-overhead2} reflects the actual burden added during real-world deployment.

\begin{table}[h]
  \centering
  \small
  \setlength{\tabcolsep}{5pt}
  \caption{Inference-time efficiency comparison for real-world deployment.
Change is computed relative to the base GO-1.}
  \begin{tabular}{lcc}
    \toprule
    Metric & GO-1 & GO-1 + EFN (ours) \\
    \midrule
    Per-step latency (ms) $\downarrow$
      & \textbf{35.7}
      & 37.2 {\color{red}{(+4.2\%)}} \\

    Peak GPU memory (MB) $\downarrow$
      & \textbf{12,851}
      & 13,464 {\color{red}{(+4.77\%)}} \\

    Avg. steps / episode $\downarrow$
      & 491.8
      & \textbf{435.1} {\color{ForestGreen}{(-11.5\%)}} \\

    Episode time (s) $\downarrow$
      & 17.6
      & \textbf{16.2} {\color{ForestGreen}{(-7.9\%)}} \\

    \bottomrule
  \end{tabular}
  \label{tab:efn-overhead2}
  \vspace{-5pt}
\end{table}

Integrating EFN increases per-step latency by only $4.2\%$ and peak GPU memory by $4.77\%$, while substantially reducing the number of steps required to complete an episode (\(-11.5\%\)). This yields a net improvement in real-world episode time (\(-7.9\%\)). Overall, EFN adds minimal system overhead yet delivers measurable efficiency gains in deployment.

\subsection{Residual correction vs. direct action prediction}
\label{app:direct-action}

For completeness, we also consider a ``direct-action'' variant of EFN where the base VLA policy is disabled and EFN is trained to directly output absolute actions. 
Concretely, instead of
\begin{equation}
    \mathbf{a}_t \;=\; \mathbf{a}^{\text{base}}_t \;+\; \Delta\mathbf{a}^{\text{EFN}}_t,
\end{equation}
we let the EFN head produce the full action
\begin{equation}
    \mathbf{a}_t \;=\; \mathbf{a}^{\text{EFN}}_t,
\end{equation}
using the same frozen VLA features and the same lightweight MLP head as in our residual formulation (see Tab.~\ref{tab:efn-latency-ablation} for the architectural footprint). 
All other training hyperparameters (SAC configuration, reward shaping, and interaction budget) are kept identical to the residual version.

\begin{table}[h]
\centering
\caption{Residual vs. direct-action EFN on the LIBERO benchmark.
The direct-action variant fails to learn and remains close to zero.}
\begin{tabular}{lcc}
\toprule
Method &  Avg. Succ \\
\midrule
OpenVLA  & 76.5 \\
OpenVLA+EFN (residual) & 87.0 \\
OpenVLA+EFN (direct-action) & \textbf{0} \\
\bottomrule
\end{tabular}
\label{zero}
\end{table}

Empirically, this direct-action EFN fails to learn meaningful behavior in our environments:
across both simulation and the real-world platform, the success rate remains close to zero throughout training (as shown in Table~\ref{zero}), and the learning curves exhibit strong oscillations rather than stable improvement. We therefore omit this variant from the main tables and report it here only as a negative result.
We hypothesize two main reasons.
First, by removing the base policy we effectively discard the generalization ability acquired by the pre-trained VLA, forcing a small EFN head to re-learn the entire control policy from sparse rewards and high-dimensional observations.
Second, the head is deliberately designed to be extremely lightweight to meet our latency budget; while this is sufficient for modeling residual corrections around a strong backbone, it appears under-parameterized for modeling the full continuous action manifold from scratch.
Taken together, these findings justify our choice of treating EFN as a residual corrector rather than a standalone policy.

\section{Visualization}
\label{vis}

To complement the quantitative results in Table~2 in the main paper, we provide real-world demonstration videos recorded on the AgiBot-G1 manipulation platform in the \texttt{demo/} directory. For each real-world task evaluated in Table~2 in the main paper, we include a video of the EFN-augmented policy executing the task (e.g., \texttt{StockLift\_with\_EFN.mp4}), illustrating how the deployed agent uses retrieved experiences to refine the base VLA policy and produce more reliable executions.

Due to space limitations in the supplementary material, we provide one paired comparison (with vs.\ without EFN) for the \textbf{most challenging} task, \emph{DrawerStore}, whose baseline success rate without EFN is only 5.3\% while EFN raises it to 58.7\% (see Table~2 in the main paper). The video \texttt{DrawerStore\_without\_EFN.mp4} was obtained after nearly twenty recording attempts, reflecting the low baseline success probability; even in this successful rollout, the robot exhibits hesitant behavior, including repeated grasps and an imprecise final release that almost leads to failure. In contrast, \texttt{DrawerStore\_with\_EFN.mp4} was recorded easily and shows a smooth, near-perfect execution of the same task, with a decisive grasp, accurate drawer interaction, and stable object placement. This qualitative gap is consistent with the quantitative improvement brought by EFN and provides an intuitive illustration of how experience feedback stabilizes behavior on the hardest real-world task.

\end{document}